\documentclass{article}

\PassOptionsToPackage{numbers, compress}{natbib}
\usepackage[preprint]{neurips_2024}

\usepackage[utf8]{inputenc} % allow utf-8 input
\usepackage[T1]{fontenc}    % use 8-bit T1 fonts
\usepackage{hyperref}       % hyperlinks
\usepackage{url}            % simple URL typesetting
\usepackage{booktabs}       % professional-quality tables
\usepackage{amsfonts}       % blackboard math symbols
\usepackage{nicefrac}       % compact symbols for 1/2, etc.
\usepackage{microtype}      % microtypography
\usepackage{xcolor}         % colors

\usepackage{epsfig}
\usepackage{wrapfig}
\usepackage{graphicx}
\usepackage{multirow}
\usepackage{makecell}
\usepackage{booktabs}
\usepackage{amssymb}
\usepackage{amsmath}
\usepackage{longtable}
\usepackage{xcolor}
\usepackage{colortbl}
\usepackage{hhline}
\usepackage{color}
\usepackage[misc]{ifsym}
\usepackage{diagbox}
\usepackage{bbding}
\usepackage{caption}
\usepackage{tabulary}
\usepackage{enumitem}

\title{Hierarchical Side-Tuning for Vision Transformers}

\author{%
  \centerline{
  Weifeng Lin$^{1}$ \quad
  Ziheng Wu$^{2,3}$ \quad
  Wentao Yang$^{1}$ \quad
  Mingxin Huang$^{1}$ } \\
  \textbf{
  Jun Huang${^3}$ \quad
  Lianwen Jin${^1}$} \vspace{0.6em}
  \\
  \centerline{
  $^{1}$South China University of Technology   \quad
  $^{2}$Zhejiang University  \quad
  $^{3}$Alibaba Group
  }
}

\begin{document}

\maketitle

\begin{abstract}
Fine-tuning pre-trained Vision Transformers (ViTs) has showcased significant promise in enhancing visual recognition tasks. Yet, the demand for individualized and comprehensive fine-tuning processes for each task entails substantial computational and memory costs, posing a considerable challenge.
Recent advancements in Parameter-Efficient Transfer Learning (PETL) have shown potential for achieving high performance with fewer parameter updates compared to full fine-tuning. 
However, their effectiveness is primarily observed in simple tasks like image classification, while they encounter challenges with more complex vision tasks like dense prediction.
To address this gap, this study aims to identify an effective tuning method that caters to a wider range of visual tasks. In this paper, we introduce Hierarchical Side-Tuning (HST), an innovative PETL method facilitating the transfer of ViT models to diverse downstream tasks.
Diverging from existing methods that focus solely on fine-tuning parameters within specific input spaces or modules, HST employs a lightweight Hierarchical Side Network (HSN). This network leverages intermediate activations from the ViT backbone to model multi-scale features, enhancing prediction capabilities.
To evaluate HST, we conducted comprehensive experiments across a range of visual tasks, including classification, object detection, instance segmentation, and semantic segmentation. Remarkably, HST achieved state-of-the-art performance in 13 out of the 19 tasks on the VTAB-1K benchmark, with the highest average Top-1 accuracy of \textbf{76.1\%}, while fine-tuning a mere \textbf{0.78M} parameters. When applied to object detection and semantic segmentation tasks on the COCO and ADE20K testdev benchmarks, HST outperformed existing PETL methods and even surpassed full fine-tuning. Code is available at \url{https://github.com/AFeng-x/HST}

\end{abstract}

\section{Introduction}
Recently, Vision Transformers (ViTs) have achieved remarkable success~\cite{vit}. Inspired by the achievements of large language models~\cite{raffel2020exploring, brown2020language, devlin2018bert}, there is a growing enthusiasm for leveraging the pre-trained knowledge embedded within ViTs, such as CLIP~\cite{radford2021learning}, MAE~\cite{he2022masked} and DINO~\cite{caron2021emerging}, to enhance performance in downstream tasks.
However, the rapid increase in model size has rendered full fine-tuning of these pre-trained models for downstream tasks impractical due to the associated storage overhead.
To tackle this challenge, many studies have introduced Parameter-Efficient Transfer Learning (PETL)~\cite{lian2022scaling,hu2021lora,jia2022visual,houlsby2019parameter} to develop a high-performing tuning system without the necessity of training an entirely new model for each task.
The PETL methods function by either selecting a subset of pre-trained parameters or introducing a constrained set of trainable parameters into the backbone, all the while maintaining the majority of the original parameters in a fixed state.

Although PETL methods have achieved considerable success, it's crucial to acknowledge their limitations when applied to broader visual tasks. Most of PETL techniques excel in image classification but struggle with more complex tasks such as dense prediction, which includes object detection and segmentation. These tasks differ fundamentally from classification as they require discernment of multi-grained features. Simply inserting a limited number of trainable parameters to the backbone often falls short in capturing multi-scale features, leading to suboptimal performance in these demanding tasks.

\begin{figure}[t]
    \centering
    \includegraphics[width=0.9\linewidth]{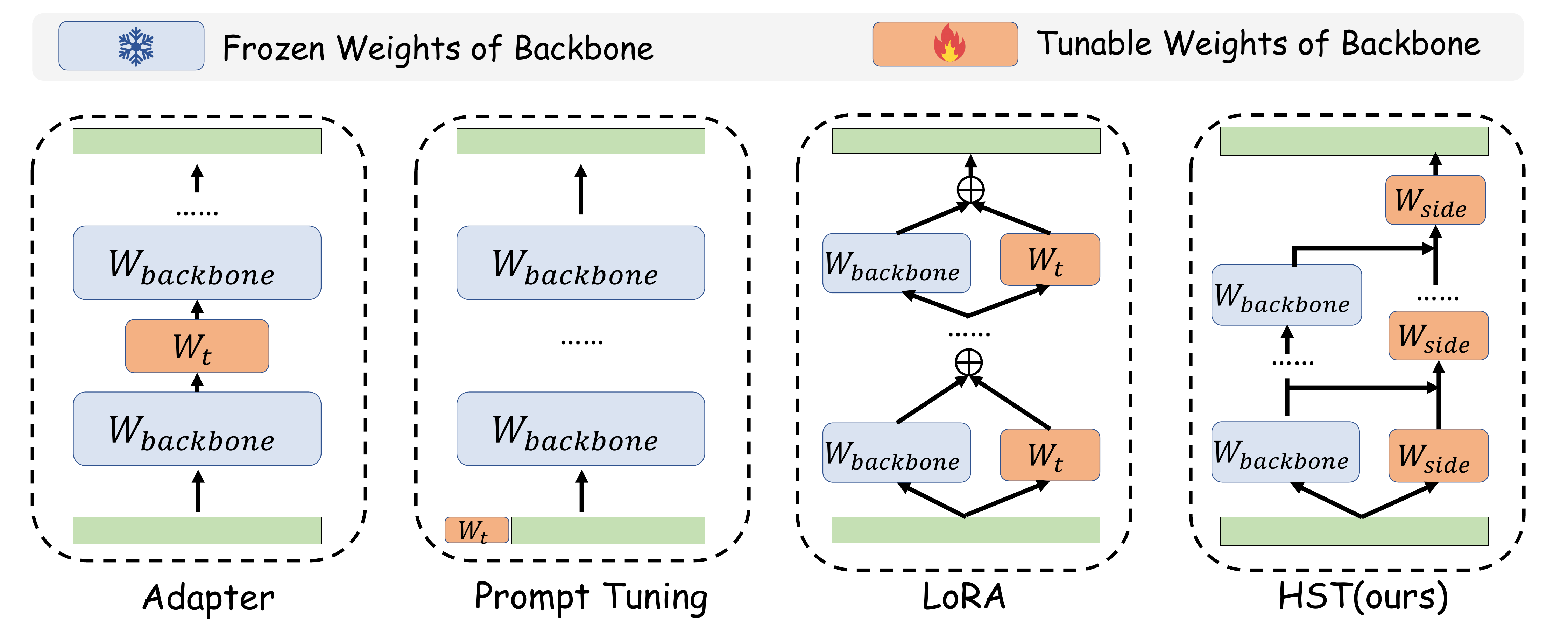}
    \caption{\textbf{Previous paradigm vs. our paradigm,} including Adapter, Prompt Tuning, LoRA and our Hierarchical Side-Tuning.}
    \label{fig:paradigm}
    \vspace{-4mm}
\end{figure}

Therefore, we propose a versatile PETL method named Hierarchical Side-Tuning (HST). As illustrated in Figure \ref{fig:paradigm}, different from other methods, we segregate most of the trainable parameters from the backbone. This partitioning facilitates the creation of a lightweight Hierarchical Side Network (HSN), proficient at modeling multi-scale features and efficiently adapting the entire model to diverse tasks.
To fully leverage the pre-trained knowledge embedded within ViT, we introduce two key modules to enhance the integration of ViT's intermediate activations: the Meta-Register and the Transformation Bridge (T-Bridge). The Meta-Register consists of one trainable token, which adapt to capture crucial global features within each Transformer block of ViT. Meanwhile, the Transformation Bridge is specifically designed to effectively bridge and preprocess the intermediate activations.
Within HSN, we develop the Side block as its foundational component. This block takes pre-trained ViT's intermediate activations and the multi-scale features of images as inputs, allowing for feature fusion based on inputs of varying granularity. 
Through the stacking of Side blocks, the proposed HSN demonstrates the capability to model multi-scale features similar to those of hierarchical ViT variants~\cite{swin,pvt}, which have been proven to be adaptable and effective in tackling a wide range of visual tasks.

We conduct comprehensive experiments on HST, spanning image classification, object detection, instance segmentation and semantic segmentation.
Overall, HST achieves state-of-the-art (SOTA) performance compared to existing PETL methods with fewer trainable parameters. When compared to the full fine-tuning method, HST exhibited a significant performance improvement of 10.5\% (76.1\% vs. 65.6\%) in terms of average Top-1 accuracy on VTAB-1K~\cite{zhai2019large}, with merely 0.78M trainable parameters. Furthermore, our HST outperformed other PETL methods by a substantial margin and achieve comparable performance to full fine-tuning method on MS COCO~\cite{lin2014microsoft} and ADE20K~\cite{zhou2017scene} testdev benchmarks for dense prediction tasks.

\section{Related Work}
\paragraph{Vision Transformer}
Transformers~\cite{vaswani2017attention} have showcased remarkable performance on Natural Language Processing(NLP) tasks. ViT~\cite{vit} is the first work to generalize the Transformer to the vision task without much modification. 
Subsequently, inspired by its vast success, various pre-training methods based on the ViT architecture have emerged, including CLIP~\cite{radford2021learning}, BEIT~\cite{bao2021beit}, MAE~\cite{he2022masked}, and DINO~\cite{caron2021emerging}, among others. These methods illustrate that adopting pre-trained Transformer models for downstream tasks can alleviate the training difficulty and lead to the swift attainment of promising results. However, as model sizes increase, the need for individualized and comprehensive fine-tuning processes for each downstream task incurs significant computational and memory costs. Therefore, addressing the challenge of adapting pre-trained ViT to downstream tasks in a manner that is both parameter and memory efficient remains a critical open issue.

\vspace{-2mm}
\paragraph{Parameter-Efficient Transfer Learning}
As model sizes continue to expand rapidly, there has been a growing focus on Parameter-Efficient Transfer Learning (PETL)~\cite{liu2022p,lester2021power,mao2021unipelt,he2022hyperprompt,he2021towards}. PETL targets re-adopting a large-scale pre-trained model as the starting point and only fine-tuning a few parameters to achieve fair performance competitive to a fully tuned one.
Adapter-based and prompt-based tuning stand as two main paradigms for pre-trained models. 
As depicted in Figure~\ref{fig:paradigm}, Visual Prompt Tuning (VPT)~\cite{jia2022visual} utilizes prompts, comprised of trainable tokens, within the input sequence of the vision Transformer. 
However, VPT necessitates a search for the optimal prompt length for each specific downstream task, a process that can be time-consuming.
Adapter~\cite{houlsby2019parameter} proposes an MLP-like module with two fully connected layers inserted into the backbone. 
Unlike injecting trainable modules into the transformer blocks, LoRA~\cite{hu2021lora} learns to optimize a low-rank decomposition matrix with a low intrinsic dimension to project the matrices of multi-head self-attention. 
Side-Tuning~\cite{zhang2020side} involves learning a side model S(x) and combining it with a pre-trained base model B(x) in the last layer, without any interaction at the intermediate feature layers.
LST~\cite{sung2022lst} was initially introduced in the field of NLP to address training efficiency issues. It involves freezing the pre-trained model and utilizing intermediate features as supplementary inputs to train a side network. However, it has not been proven to be effective in vision models and initializing the side network poses a challenge.

\vspace{-2mm}
\paragraph{Decoders for ViT}
ViT is a powerful alternative to standard ConvNets for image classification. However, the original ViT is a plain, non-hierarchical architecture. As a result, it cannot be relatively straightforward to replace a ConvNet with the backbone for dense prediction. 
Recently, UViT~\cite{chen2021simple} uses single-scale feature maps for the detector heads, which modifies the architecture during pre-training. Unlike UViT, several studies~\cite{li2021benchmarking,li2022exploring} focus on using multi-scale adaptor to maintain the task-agnostic nature of the backbone. 
Furthermore, SETR~\cite{zheng2021rethinking} develops several CNN decoders for semantic segmentation. 
Vit-Adapter~\cite{vit_adapter} design several modules and operations to reorganize multi-scale features for dense prediction. However, it primarily focus on enhancing ViT's performance by employing full fine-tuning. In the current era of large-scale models, conducting full fine-tuning for each downstream task has become increasingly challenging and requires substantial storage space. Thus, the challenge persists in enhancing performance of dense prediction under parameter-efficient fine-tuning and our work is dedicated to addressing this challenge.

\vspace{-2mm}
\section{Hierarchical Side-Tuning}
\begin{figure}[h]
\vspace{-2mm}
    \centering
    \includegraphics[width=0.95\linewidth]{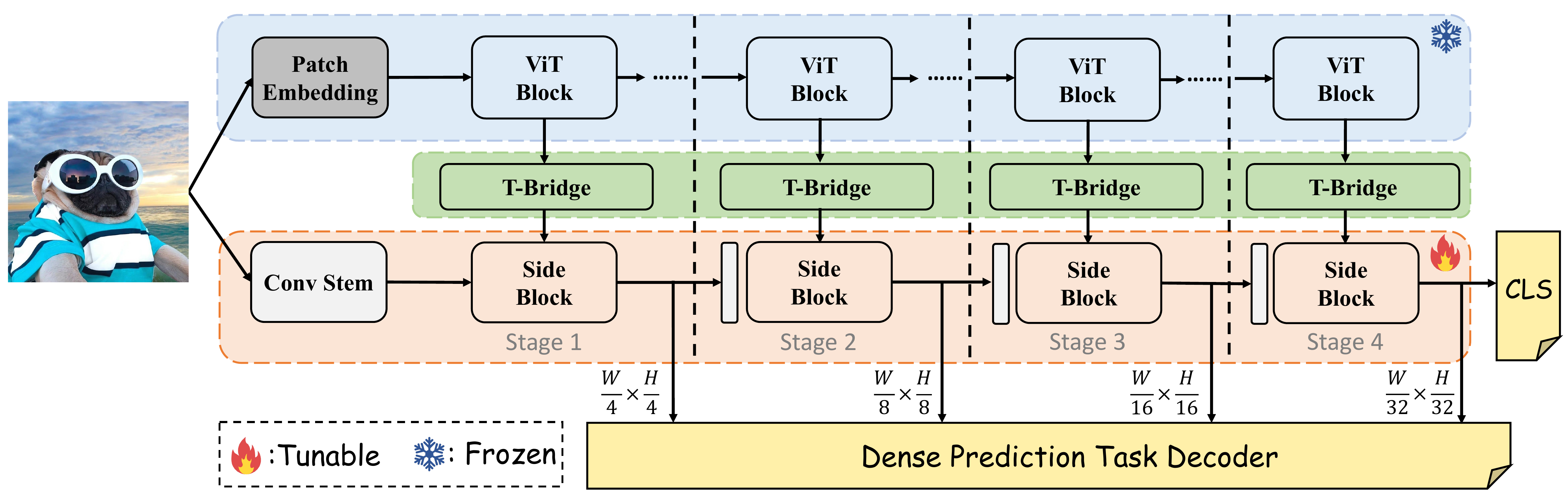}
    \caption{
    \textbf{Overall architecture of HST.} The \textcolor{blue}{Blue Section} represents the plain ViT, with its weights kept frozen. The \textcolor{green!60!black}{Green Section} is referred to as the Transformation Bridge (T-Bridge).
    The \textcolor{pink}{Pink Section} is the proposed Hierarchical Side Network (HSN), composed of a convolutional stem followed by a sequence of $L$ Side blocks. 
    }
\label{fig:arch}
\vspace{-4mm}
\end{figure}

\subsection{Overview}
As illustrated in Figure~\ref{fig:arch}, the HST architecture comprises two essential components: the Transformation Bridge (T-Bridge) and the Hierarchical Side Network (HSN). The HSN is bulit to receive and integrating multi-scale features extracted from the input image, along with intermediate activations from the pre-trained ViT. It is structured into four stages, each with downsampling rates of $\{4, 8, 16, 32\}$, responsible for generating feature pyramids at various resolutions. These pyramids are then efficiently connected with downstream task decoders. Notably, we align the number of Side blocks with the number of ViT's blocks and evenly distribute them across these four stages.
The T-Bridge plays the role of facilitating the seamless integration of intermediate activations derived from the ViT into the HSN. Additionally, within the ViT backbone, we introduce the Meta-Register, leveraging it to extract essential task-specific feature information from every Transformer block.

\subsection{Meta-Register}

\begin{figure}[t]
\vspace{-2mm}
    \centering
    \includegraphics[width=0.88\linewidth]{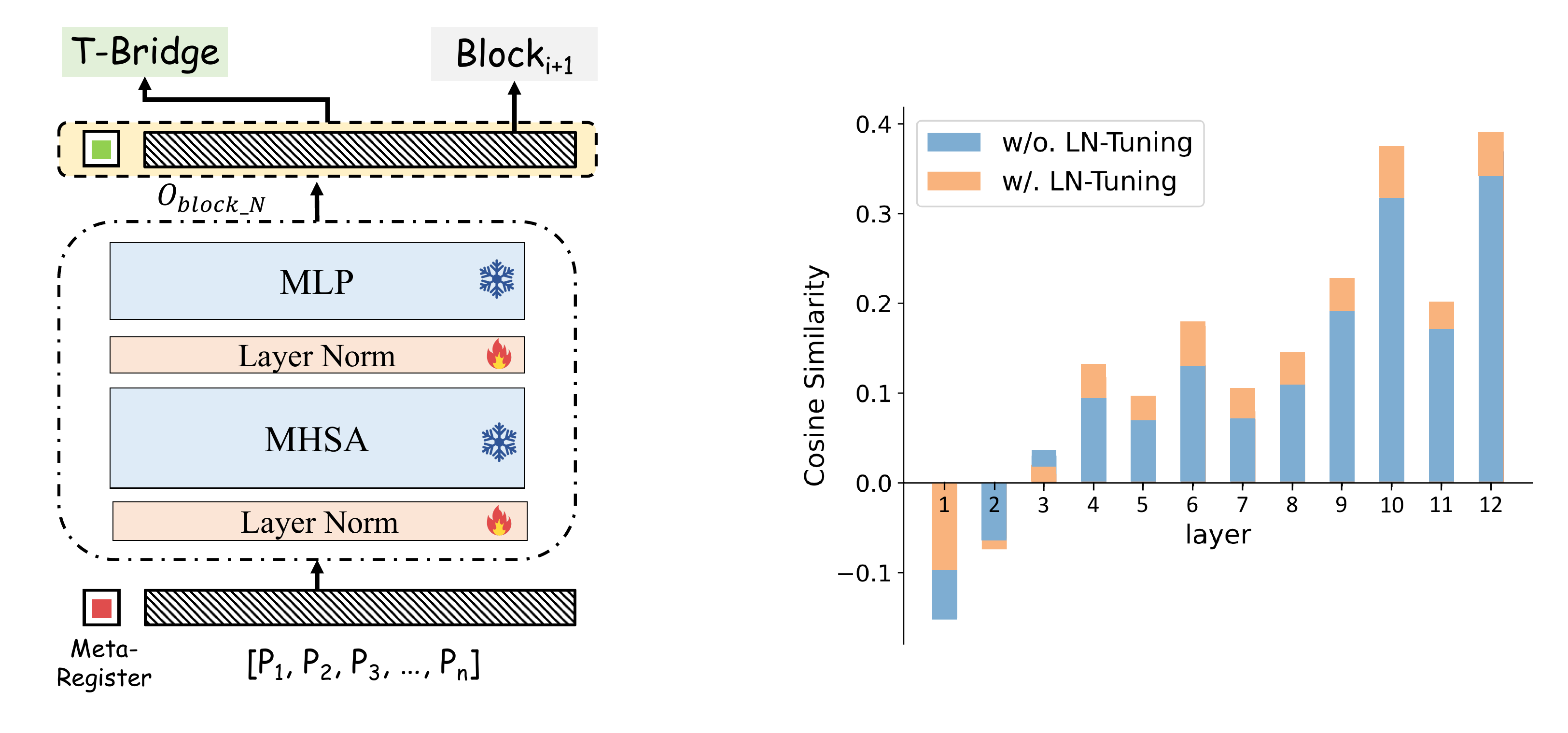}
    \vspace{-2mm}
    \caption{
    \textbf{Left: }Meta-Register and layer norm tuning. \textbf{Right: }Comparisons of cosine similarity between the output features of Meta-Register and input image tokens.
    }
\label{fig:Meta-Register}
\vspace{-2mm}
\end{figure}

Existing prompt-based tuning techniques~\cite{jia2022visual,lester2021power,li2021prefix} have two significant limitations: $(i)$ They rely on manual selection to determine the optimal prompt length for each task, and sometimes the number of prompts can even extend to several hundred, placing a substantial burden on both training and inference. $(ii)$ The output features of prompts are discarded after passing through the Transformer layer, resulting in the underutilization of valuable learning information contained within the prompts. 
Conversely, in our study, we introduce the Meta-Register, which comprises a few trainable tokens. Unlike existing prompt-based tuning techniques, we require only one trainable token in the Meta-Register, tasked with capturing crucial global features within each Transformer block. Furthermore, we input the features of the Meta-Register into the Transformation Bridge as intermediate activations, alongside the features of image tokens.
However, we have observed that the distribution of the Meta-Register differs from that of the image tokens. This disparity hampers our ability to effectively model them within the side network we've constructed. To address this issue, we propose unfreezing the weights of the Layer Normalization (LN) layer within the Transformer block. Tuning the LN layers can efficiently alter the mean and variance of the feature distribution, thereby aiding in preserving the relative magnitudes among different features within the same sample.
Figure~\ref{fig:Meta-Register} illustrates the cosine similarity between the output features of Meta-Register and the image tokens in each Transformer layer. It is evident that, with LN tuning, the Meta-Register progressively aligns more closely with the vector direction of the image tokens across layers. This alignment enables us to effectively leverage the output features of the Meta-Register in the Transformation Bridge and Side blocks. It is worth noting that training the layer normalization layers adds less than 0.1M trainable parameters, while not incurring additional training resource overhead, which is a simple yet important strategy.

\begin{wrapfigure}{r}{0.38\textwidth}
    \vspace{-2mm}
    \includegraphics[width=0.95\linewidth]{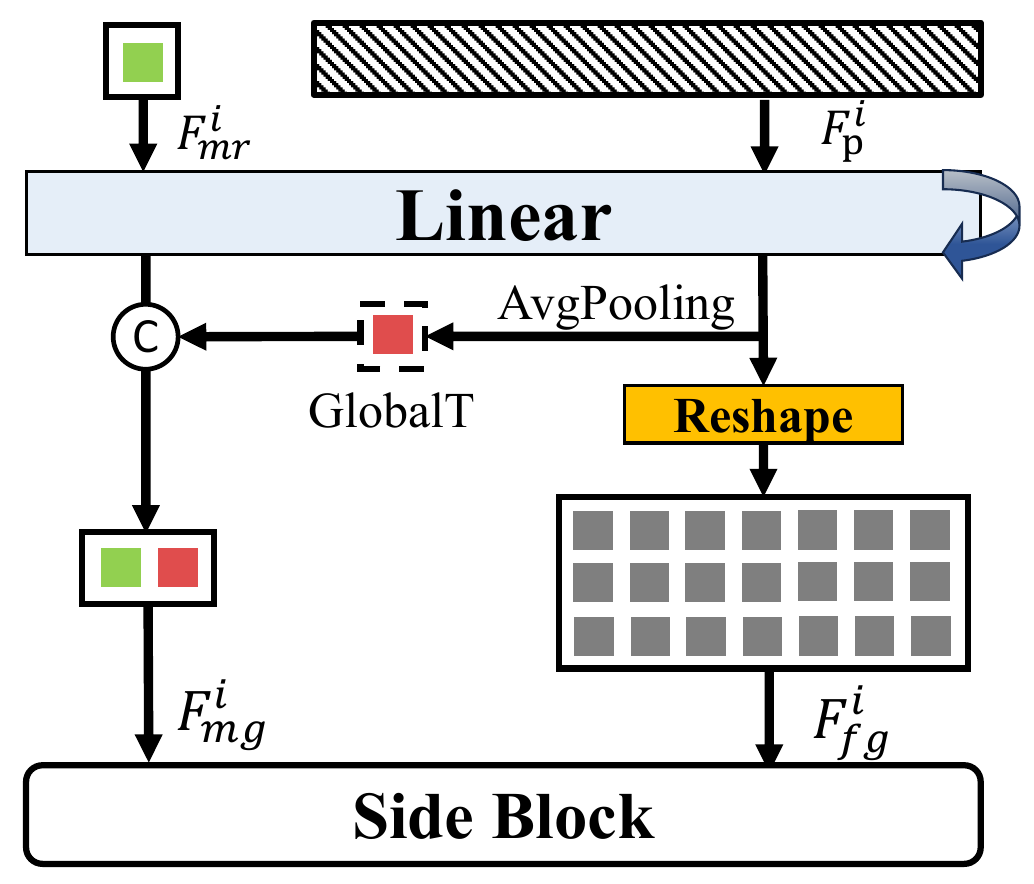}
    \caption{Transformation Bridge.}
    \label{fig:TB}
    \vspace{-6mm}
\end{wrapfigure}

\subsection{Transformation Bridge}
Given the discrepancy in shapes and dimensions between the intermediate activations derived from ViT and the multi-scale features within the Hierarchical Side Network (HSN), direct injection becomes unfeasible. Hence, we introduce a mid-processing module named the Transformation Bridge (T-Bridge), which consists of two pivotal operations: Dual-Branch Separation and Linear Weight Sharing.

\paragraph{Dual-Branch Separation}
As depicted in Figure~\ref{fig:TB}, the features of the Meta-Register $\mathcal{F}^{i}_{mr}$ and image tokens $\mathcal{F}^{i}_{p}$ initially undergo transformation through a linear layer to ensure alignment with the various stages within the HSN. Subsequently, we divide the features into two distinct branches: the Meta-Global branch and the Fine-Grained Branch. 
To enhance the global information in the first branch, we average all image tokens to yield a single token, named 'GlobalT', which is then concatenated with the Meta-Register to form the Meta-Global branches $\mathcal{F}^{i}_{mg}$. 
The Fine-Grained branch $\mathcal{F}^{i}_{fg}$ utilizes bilinear interpolation $\mathcal{T}$ to reshape the image tokens. This reshaping operation aligns the resolution with that of the corresponding stage's feature within the HSN.
The whole process can be formulated as follows:
\begin{equation}
\mathcal{F}^{i}_{mg} = [W_{j}\mathcal{F}^{i}_{mr}, AvgPooling(W_{j}\mathcal{F}^{i}_{p})]
\end{equation}
\begin{equation}
\mathcal{F}^{i}_{fg} = \mathcal{T}({W_{j}\mathcal{F}^{i}_{p}})
\end{equation}
where $i$ denotes $i$-th ViT block's output, and $W_{j}$ is the weight matrices of linear layer in $j$-th stage. 

\paragraph{Linear Weight Sharing}
We propose to share the weight of linear layer in T-Bridge for different intermediate features.
Specifically, every T-Bridge within the same stage share a common linear layer.
This approach offers the distinct advantage of reducing the number of trainable parameters. Simultaneously, it enables information interaction within the same stage, thereby achieving effects comparable to those obtained with multiple linear layers.

\subsection{Side Block}
In this section, we detail the proposed Side block that forms the fundamental building block of HSN construction. The Side block comprises a cross-attention layer and a feed-forward network (FFN), which collectively empower the modeling of intermediate features from pre-trained model and multi-scale features.
Considering the unique characteristics of the two input branches, we introduce them into the Side block through distinct approaches, specifically termed Meta-Global Injection and Fine-Grained Injection.
\begin{figure}[h]
    \centering
    \includegraphics[width=0.95\linewidth]{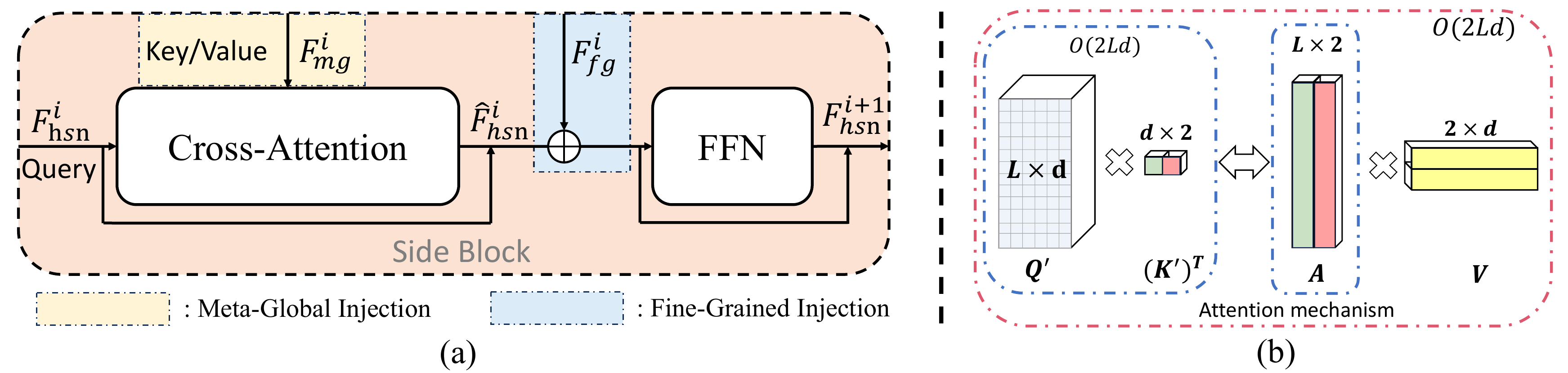}
    \vspace{-2mm}
    \caption{
    \textbf{Side Block.} (a) The schematic illustration of the proposed Side Block. (b) Illustration of linear complexity of cross-attention in Side block.
    }
    \vspace{-2mm}
    \label{fig:LCA}
\end{figure}

% \subsubsection{Meta-Global Injection}
\noindent \textbf{Meta-Global Injection. }
As illustrated in Figure~\ref{fig:LCA}(a), we utilize the multi-scale feature from HSN as the $query (Q)$  matrix and employ Meta-Global tokens as the $key (K)$ and $value (V)$ matrices for performing cross attention. This process is defined as follows:
\begin{equation}
\textstyle{
    ( (Q_{hsn}) ({K_{mg}})^T) V_{mg} = A V_{mg}
    \label{eq:linear}
}\end{equation}
where $Q_{hsn} \in \mathbb{R}^{L \times d}$, $(K_{mg})^T \in \mathbb{R}^{d \times 2}$, and $V_{mg} \in \mathbb{R}^{2 \times d}$. Here, $L$ denotes the length of the multi-scale input sequence and $d$ signifies the feature dimension.
This approach provides us with the advantage of a computation complexity of $O(2Ld)$. Notably, $d$ is significantly smaller than the input sequence length. This allows us to effectively inject global priors into the side network while also reducing the computational complexity of attention to linear, significantly improving the training and inference efficiency of the HSN.

\vspace{1mm}
\noindent \textbf{Fine-Grained Injection. }
After cross-attention, we obtain the output feature $\mathaccent"705E{F}^{i}_{hsn}$ , which can be written as follows:
\begin{equation}
	\mathcal{\hat{F}}^{i}_{hsn}=\mathcal{F}^{i}_{hsn} + {\rm Cross Attention}(\mathcal{F}^{i}_{hsn}, \mathcal{F}^{i}_{mg}),
\label{injection}
\end{equation}
where $i$ denotes $i$-th block in HST and ViT. Next, we incorporate the fine-grained branch ${F}^{i}_{fg}$ into the Side block. Specifically, we perform an element-wise addition of the obtained $\mathaccent"705E{F}^{i}_{hsn}$ and ${F}^{i}_{fg}$ after the cross-attention layer. Subsequently, a feed-forward network (FFN) is applied for further feature modeling. This procedure can be represented as follows:
\begin{equation}
    {F}^{i+1}_{hsn}=\mathaccent"705E{F}^{i}_{hsn} + {F}^{i}_{fg} + {\rm FFN}(\mathaccent"705E{F}^{i}_{hsn} + {F}^{i}_{fg})
    \label{cffn}
\end{equation}
where the generated feature ${F}^{i+1}_{hsn}$ will be used as the input of the next Side block.

\section{Experiments}
\subsection{Experimental Settings}\label{ex_settings}
\paragraph{Detailed Architectures Specifications} As shown in Table~\ref{tab:arch_setting}, HSN's architecture varies the dimensions and attention heads across stages, increasing with layer depth. In classification experiments, HSN's dimensions are significantly smaller than ViT's (768). For dense prediction tasks, we choose slightly larger dimensions to ensure sufficient capacity for handling dense prediction tasks. Notably, neck modules like FPN also adopt dimensions of [64, 128, 256, 384], which sets them apart from other methods where neck modules maintain ViT's dimensions, thus requiring fewer training parameters.

\begin{table}[h]\small
    \centering
    \renewcommand{\arraystretch}{1.0}
	\setlength\tabcolsep{0.6mm}{
	\begin{tabular}{l|ccc|ccc|c}
		\toprule
	    \multirow{2}{*}{Model Size (Task)}  & \multicolumn{3}{c|}{ViT}   & \multicolumn{3}{c|}{HSN} & \#Trainable  \\
	   & Embed.Dims & Depth & Attn.Heads  & Embed.Dims & Attn.Heads & Depths & Params \\
		\midrule
        % ViT-B/HSN-S (Cls) & 768 & 12 & 12 & {[}16,32,48,64{]} & \multirow{5}{*}{[2,4,8,12]} & {[}3,3,3,3{]} & 0.28M \\
        ViT-B/HSN-B (Cls) & 768 & 12 & 12 & [32,48,64,72] & \multirow{4}{*}{[2,4,8,12]} & {[}3,3,3,3{]} & 0.78M \\
        ViT-B/HSN-L (Det/Seg) & 768 & 12 & 12 & [64,128,256,384] & ~ & {[}3,3,3,3{]} & 13.21M \\
        ViT-L/HSN-B (Cls) & 1024 & 24 & 16 & [32,48,64,72] & ~ & [6,6,6,6] & 0.78M \\
        ViT-L/HSN-L (Det/Seg) & 1024 & 24 & 16 & [64,128,256,384] & ~ & {[}6,6,6,6{]} & 19.86M \\
	\bottomrule
	\end{tabular}}
        \vspace{1mm}
	\caption{Detailed architectures specifications.}
	\label{tab:arch_setting}
\vspace{-4mm}
\end{table}

\paragraph{Pre-trained backbone}
To ensure fair comparisons, we adopt the plain Vision Transformer (ViT)~\cite{vit} pre-trained on ImageNet-21K~\citep{deng2009imagenet} and MAE~\cite{he2022masked} as the initialization for fine-tuning on downstream tasks.

\vspace{-2mm}
\paragraph{Downstream tasks}
We evaluate the performance of HST on both image classification and dense prediction tasks to confirm its effectiveness. 
Due to ViT producing feature maps at a single scale (e.g., 1/16th), it could not be adapted to work with a feature pyramid network (FPN)~\citep{lin2017feature}. Therefore, we follow~\citep{li2021benchmarking} to either upsample or downsample intermediate ViT feature maps by placing four resolution-modifying modules to adapt the single-scale ViT to the multi-scale FPN.
In this way, similar to recognition tasks, we only need to train the newly added parameters and specific-task head, enabling us to achieve parameter-efficient transfer learning for dense prediction tasks. 
% We provide detailed description in the Appendix~\ref{app:detailed_datasets}.

\subsection{Performance Comparisons on Image Classification}\label{cls}
\begin{table*}[h]
        % \vspace{-4mm}
	\setlength\tabcolsep{2.4pt}
	\centering
	%\vspace{-0.3cm}
	\scalebox{0.7}{
            \begin{tabular}{c|ccccccc|cccc|cccccccc|cc}
		\toprule
			& & \multicolumn{6}{c|}{\textbf{Natural}} & \multicolumn{4}{c|}{\textbf{Specialized}} & \multicolumn{8}{c|} {\textbf{Structured}} &  &   \\ 
            \midrule
			Method & \rotatebox{90}{CIFAR-100} & \rotatebox{90}{Caltech101} & \rotatebox{90}{DTD} & \rotatebox{90}{Flowers102} & \rotatebox{90}{Pets} & \rotatebox{90}{SVHN}  & \rotatebox{90}{Sun397} & \rotatebox{90}{Camelyon~} & \rotatebox{90}{EuroSAT}   & \rotatebox{90}{Resisc45}  & \rotatebox{90}{Retinopathy} & \rotatebox{90}{Clevr/count} & \rotatebox{90}{Clevr/distance}  & \rotatebox{90}{DMLab} & \rotatebox{90}{KITTI/distance~}  & \rotatebox{90}{dSprites/loc} & \rotatebox{90}{dSprites/ori}   & \rotatebox{90}{SmallNORB/azi~}  & \rotatebox{90}{SmallNORB/ele~} & \rotatebox{90}{Average (\%)} & \rotatebox{90}{Params. (M)}    \\
		\midrule
			
            Full fine-tuning~\cite{jia2022visual}  & 68.9 & 87.7 & 64.3 & 97.2  & 86.9 & 87.4 & 38.8 & 79.7    & 95.7   & 84.2         & 73.9           & 56.3          & 58.6          & 41.7 & 65.5 & 57.5         & 46.7         & 25.7           & 29.1  &  65.57  & 85.84   \\ 
			
            Linear probing~\cite{jia2022visual}    & 63.4  & 85.0  & 63.2 & 97.0 & 86.3 & 36.6 & 51.0 & 78.5 & 87.5 & 68.6 & 74.0 & 34.3 & 30.6 & 33.2 & 55.4 & 12.5 & 20.0 & 9.6 & 19.2 & 52.94 & 0.04 \\
		\midrule
			
            Adapter~\cite{houlsby2019parameter} & 74.1 &  86.1 & 63.2 & 97.7 & 87.0 & 34.6 & 50.8 & 76.3 & 88.0 & 73.1 & 70.5 & 45.7 & 37.4 & 31.2 & 53.2 & 30.3 & 25.4 & 13.8 & 22.1 & 55.82 & 0.27 \\

            % Sidetune~\cite{zhang2020side} & 46.88 & - & 60.7 & 60.8 & 53.6 & 95.5 & 66.7& 34.9& 35.3 & 58.5 & 87.7& 65.2& 61.0& 27.6& 22.6& 31.3& 51.7 & 8.2& 14.4& 9.8& 21.8 \\
        
            Bias~\cite{zaken2021bitfit} &  72.8 & 87.0 & 59.2 & 97.5 & 85.3 & 59.9 & 51.4 & 78.7& 91.6& 72.9& 69.8& 61.5 & 55.6&32.4 & 55.9& 66.6& 40.0& 15.7& 25.1 & 62.05 & \textbf{0.14}  \\
			
            % VPT-Shallow~\cite{jia2022visual} & 77.7 & 86.9 & 62.6& 97.5& 87.3& 74.5& 51.2& 78.2& 92.0& 75.6& 72.9& 50.5& 58.6& 40.5& 67.1 & 68.7& 36.1& 20.2& 34.1 & 64.85 & \textbf{0.11} \\
			
            VPT-Deep~\cite{jia2022visual} & \underline{78.8} & 90.8 & \underline{65.8} & 98.0 & 88.3 & 78.1& 49.6& 81.8& 96.1 & 83.4& 68.4& 68.5 & 60.0 & 46.5 & 72.8 & 73.6 & 47.9 & \underline{32.9} & 37.8 & 69.43 & 0.60 \\

            LoRA~\cite{hu2021lora} & 67.1 & 91.4 & 69.4 & 98.8 & 90.4 & 85.3& \underline{54.0} & 84.9 & 95.3& 84.4& 73.6& \underline{82.9} & \textbf{69.2} & 49.8 & 78.5 & 75.7 & 47.1 & 31.0 & 44.0 & 72.25 & 0.29 \\

            NOAH~\cite{zhang2022neural} & 69.6 & 92.7 & 70.2 & 99.1 & 90.4 & 86.1 & 53.7 & 84.4 & 95.4 & 83.9 & 75.8 & 82.8 & \underline{68.9} & 49.9 & 81.7 & \underline{81.8} & 48.3 & 32.8 & \underline{44.2} & 73.20 & 0.36 \\

            AdaptFormer-64~\cite{chen2022adaptformer} & 70.6 & \underline{92.9} & 72.2 & \textbf{99.6} & \underline{91.3} & 86.9& \textbf{55.4} & \textbf{88.5}& \textbf{96.6}& 87.1& \textbf{76.9} & 78.5 & 62.1 & 51.9 & 81.2 & 74.6 & 52.5 & 31.5 & 39.4 & 73.10 & 1.26 \\

            SSF~\cite{lian2022scaling} &  69.0 & 92.6 & \underline{75.1} &  \underline{99.4} & \textbf{91.8} & \underline{90.2} & 52.9 & \underline{87.4} & 95.9 & \underline{87.4} & 75.5 & 75.9 & 62.3 & \underline{53.3} & 80.6 & 77.3 & \underline{54.9} & 29.5 & 37.9 & 73.10  & 0.24   \\

            % TOAST~\cite{shi2023refocusing} 
            % & \textbf{82.1} & 90.5 & 70.5 & 98.7 & 89.7 & 71.9 & 53.3 & 84.3 & 95.5 & 85.5 & 74.2 & 75.4 & 60.8 & 44.7 & 77.5 & 73.9 & 47.5 & 24.5 & 33.7 & 70.20 & 14.0 \\

            % EXPRES~\cite{das2023learning} & 
            %  78.0 & 89.6 & 68.8 & 98.7 & 88.9 & 81.9 & 51.9 & 84.8 & 96.2 & 80.9 & 74.2 & 66.5 & 60.4 & 46.5 & 77.6 & 78.0 & 49.5 & 26.1 & 35.3 & 72.90 & - \\
		
            \midrule
            HST-S \textbf{(ours)} & 76.2 & 94.8 & 74.2 & 99.6 &	90.1 &	90.8 &	47.2 &	87.8 &	96.0 &	87.0 &	75.9 &	83.8 &	61.8 &	53.9 &	\textbf{83.2} &	86.3 &	55.4 &	30.2 &	46.2 & \underline{74.75}  & 0.28  \\
            
            HST-B \textbf{(ours)} & 76.7 & \textbf{95.1} & \textbf{75.2} &	\textbf{99.6} &	91.1 &	\textbf{91.2} &	52.3 &	87.1 &	\textbf{96.6} &	\textbf{88.6} &	\underline{76.5} &	\textbf{85.4} &	63.7 &	\textbf{53.4} &	\underline{81.8} &	\textbf{87.2} &	\textbf{56.8} &	\textbf{35.8} &	\textbf{52.1} & \textbf{76.12}  & 0.78  \\
		
            \bottomrule
		\end{tabular}
	}
        \vspace{-2mm}
	\caption{Performance comparisons on the VTAB-1k benchmark with ViT-B/16 models.}
	\label{table: vtab}
	\vspace{-2mm}
\end{table*}

\paragraph{VTAB-1K Benchmark}
In Table~\ref{table: vtab}, we compare HST to other PETL methods using ViT-B/16 pre-trained on ImageNet-21K on all three splits of the VTAB-1k dataset. The results show that even with a relatively low number of trainable parameters (0.28M), HST achieves an impressive average accuracy of 74.75\%, surpassing all other methods. Moreover, as the number of trainable parameters increases to 0.78M, HST's performance improves significantly to 76.12\%.
Remarkably, HST outperforms full fine-tuning on all 19 tasks, requiring only an additional 0.9\% of the backbone parameters. Compared to SSF, LoRA, AdaptFormer, and NOAH, HST demonstrates superior performance with improvements of +3.0\%, +3.85\%, +3.0\%, and +2.9\%, respectively. Notably, substantial gains of +6.9\%, +5.4\%, and +7.9\% on Clever/Count, dSprites/loc, and SmallNORB/ele highlight the remarkable effectiveness and parameter efficiency of HST.

\begin{table*}[h]
	\centering
	\setlength{\tabcolsep}{4pt}
	\scalebox{0.9}{\begin{tabular}{c|c|ccccc}
			\toprule
			\diagbox{Method}{Dataset}&\makecell[c]{Cifar-100} & \makecell[c]{~CUB-200~ \\ -2011} & \makecell[c]{~Oxford~ \\ Flowers}  & \makecell[c]{~Stanford~ \\ Dogs}  & \makecell[c]{~Stanford~ \\ Cars}  & \makecell[c]{~Params.(M)}   \tabularnewline \midrule
            \rowcolor{gray!15} 
			Full fine-tuning & 93.8 / 88.9 & 87.3 / 83.0 & 98.8 / 90.9 & 89.4 / 84.6  & 84.5 / 91.5 & 85.98   \tabularnewline %\midrule
			Linear probing & 88.7 / 36.9 & 85.3 / 31.7 & 97.9 / 46.0 & 86.2 / 53.2 &  51.3 / 32.8 &  0.18 \tabularnewline  \midrule
			Adapter \cite{houlsby2019parameter} & 93.3 / 74.9 & 87.1 / 74.0 & 98.5 / 85.0 & 89.8 / 78.4 & 68.6 / 72.5  &  0.41 \tabularnewline 
			Bias \cite{zaken2021bitfit} & 93.4 / 76.3 & 88.4 / 74.3 & 98.8 / 84.4 & \textbf{91.2} / 80.8 & 79.4 / 73.8 & 0.28  \tabularnewline %\midrule
			VPT-Shallow \cite{jia2022visual} & 90.4 / 73.1 & 86.7 / 71.1 & 98.4 / 86.5 & 90.7 / 68.8 & 68.7 / 79.0 & 0.25 \tabularnewline %\midrule
			VPT-Deep \cite{jia2022visual} & 93.2 / 74.2 & 88.5 / 73.3 & 99.0 / 87.4 & 90.2 / 71.5 & 83.6 / 81.9 & 0.85 \tabularnewline \midrule
			HST \textbf{(ours)}   & \textbf{93.6} / \textbf{79.7} & \textbf{89.2} / \textbf{78.7} & \textbf{99.6} / \textbf{91.2} & 89.5 / \textbf{86.4} & \textbf{88.2} / \textbf{83.7}  & 0.78 \tabularnewline
			\bottomrule
		\end{tabular}
	}
        \vspace{1mm}
	\caption{Performance comparisons on CIFAR-100 and four FGVC datasets with ViT-B/16 models pre-trained on \textbf{ImageNet-21K / MAE}.}
	\label{tab:fgvc}
 \vspace{-2mm}
\end{table*}

\paragraph{General Image Benchmark}
Following VPT~\cite{jia2022visual}, we utilize four Fine-Grained Visual Classification (FGVC) datasets~\cite{wah2011caltech, nilsback2008automated, khosla2011novel, gebru2017fine} to assess the performance of our proposed HST approach. 
Additionally, we employ the CIFAR-100~\cite{krizhevsky2009learning} dataset as a general image classification benchmark to further confirm the effectiveness of HST.
To evaluate the adaptability of these PETL methods across various pre-training techniques, we predominantly choose the ViT-B/16~\cite{vit} model, pre-trained on ImageNet-21K\footnote{~\url{https://github.com/rwightman/pytorch-image-models/releases/download/v0.1-vitjx/jx_vit_base_patch16_224_in21k-e5005f0a.pth}}, and MAE\footnote{~\url{https://dl.fbaipublicfiles.com/mae/pretrain/mae_pretrain_vit_base.pth}.}~\cite{he2022masked} as the initialization for fine-tuning.
As the results shown in Table~\ref{tab:fgvc},
under ImageNet-21K pre-training, HST achieves comparable performance on the CIFAR-100 dataset (93.6\% vs. 93.8\%) and surpasses full fine-tuning on four FGVC datasets with only 0.78M trainable parameters.
In the case of MAE pre-training, it is evident that other PETL methods exhibit subpar performance, with most of them significantly falling below the level of full fine-tuning. This indicates their limited adaptability across different pre-training methods. In contrast, HST not only outperforms full fine-tuning on certain datasets but also maintains a minimal performance gap on others. This underscores the versatility and effectiveness of HST across a wide range of pre-training approaches.

\begin{table*}[h]\small
	\centering
	\renewcommand{\arraystretch}{0.5}
    \setlength\tabcolsep{0.1mm}{
    \begin{tabular}{c|c|c|cccccc|cccccc}
        \toprule
        % \hline
        \multirow{2}{*}{Backbone} & \multirow{2}{*}{Method} & \#Param & 
        \multicolumn{6}{c|}{Mask R-CNN 1$\times$ schedule} & \multicolumn{6}{c}{Mask R-CNN 3$\times$+MS schedule}\\

        ~ & ~ & (M)
        & $\rm AP^b$ & $\rm AP^b_{50}$ & $\rm AP^b_{75}$ & $\rm AP^m$ & $\rm AP^m_{50}$ & $\rm AP^m_{75}$ 
        & $\rm AP^b$ & $\rm AP^b_{50}$ & $\rm AP^b_{75}$ & $\rm AP^m$ & $\rm AP^m_{50}$ & $\rm AP^m_{75}$   \\
	\midrule
        \rowcolor{gray!15} 
        \cellcolor{white} \multirow{10}{*}{ViT-B} & Full fine-tuning &
        113.6 & 43.1 & 65.9 & 46.8 & 39.5 & 62.9 & 42.1 & 45.1 &67.2 & 48.9 & 40.5 & 63.9 & 43.0 \\
        ~ & Linear probing& 
        27.8 & 22.1 & 43.5 & 20.0 & 22.6 & 41.1 & 22.1 & 25.0 & 47.3 & 23.9 & 24.9 & 44.9 & 24.6 \\
        % \midrule
        ~ & VPT-deep~\cite{jia2022visual} & 
        28.4 & 31.1 & 55.0 & 31.1 & 30.5 & 52.0 & 31.1 & 33.4 & 57.4 & 34.3 & 32.2 & 54.0 & 33.3  \\
        ~ & AdaptFormer~\cite{chen2022adaptformer} & 
        29.0 & 32.8 & 57.4 & 33.4 & 32.2 & 54.3 & 33.1 & 36.7 & 61.6 & 38.5 & 35.1 & 58.1 & 36.6  \\
        ~ & SSF~\cite{lian2022scaling} & 
        28.0 & 35.6 & 60.2 & 37.4 & 34.4 & 57.0 & 36.0 & 36.5 & 60.6 & 38.4 & 34.8 & 57.6 & 36.3  \\
        ~ & LoRA-32~\cite{hu2021lora} & 
        28.4 & 36.2 & 60.9 & 37.5 & 35.0 & 57.9 & 36.5 & 39.3 & 64.1 & 41.6 & 37.1 & 60.6 & 39.1 \\
        ~ & HST \textbf{(ours)} & 
        30.6 & \bf 40.3 & \bf 64.3 & \bf 43.1 & \bf 38.0 & \bf 61.1 & \bf 40.0 & \bf 43.9 & \bf 67.0 & \bf 47.7 & \bf 40.4 & \bf 64.0 & \bf 43.1  \\
        \midrule
        \rowcolor{gray!15} 
        \cellcolor{white} \multirow{6}{*}{ViT-L} & Full-tuning~\cite{jia2022visual} & 
        337.3 &	45.7 &	68.9 &	49.4 &	41.5 &	65.6 &	44.6 & - & - & - & - & - & -  \\
        ~ & Linear probing~\cite{jia2022visual} & 
        33.6 &	31.6 &	56.4 &	32.0 &	31.3 &	53.3 &	32.5 & - & - & - & - & - & -  \\
        ~ & LoRA-64~\cite{hu2021lora} & 
        39.84 &	45.0 &	68.9 &	49.1 &	41.2 &	65.3 &	44.0 & - & - & - & - & - & -  \\
        ~ & HST \textbf{(ours)} & 
        39.62 & \bf	45.5 & \bf	69.0 & \bf 49.1 & \bf 41.5 & \bf 65.5 & \bf	44.3 & - & - & - & - & - & - \\
        \bottomrule
    \end{tabular}}
    \vspace{-8mm}
    \label{tab:results_detection_mask}
\end{table*}
\begin{table*}[h]\small
        \hspace{-2.5mm}
	\centering
	\renewcommand{\arraystretch}{0.93}
    \setlength\tabcolsep{0.70mm}{
    \begin{tabular}{c|c|ccccccc|cccc}
        \toprule
        % \hline
        \multirow{2}{*}{Backbone} & \multirow{2}{*}{Method} & 
        \multicolumn{7}{c|}{Cascade Mask R-CNN 3$\times$ +MS} & \multicolumn{4}{c}{ATSS 3$\times$+MS}\\
        % \cline{3-14} 
        ~ & ~
        & \#Param & $\rm AP^b$ & $\rm AP^b_{50}$ & $\rm AP^b_{75}$ & $\rm AP^m$ & $\rm AP^m_{50}$ & $\rm AP^m_{75}$ 
        & \#Param & $\rm AP^b$ & $\rm AP^b_{50}$ & $\rm AP^b_{75}$   \\
        \rowcolor{gray!15} 
        \midrule
        \cellcolor{white} \multirow{7}{*}{ViT-B} & Full fine-tuning &
        151.4M & 48.7 & 68.1 & 52.3 & 42.2 & 65.1 & 45.4 & 101.3M & 46.7 & 67.2 & 50.1 \\
        ~ & Linear probing & 
        65.6M & 35.9 & 55.3 & 38.5 & 31.4 & 52.2 & 32.2 & 15.6M & 26.0 & 43.9 & 26.4  \\
        ~ & VPT-deep~\cite{jia2022visual} & 
        66.2M & 42.2 & 62.1 & 45.4 & 37.1 & 59.2 & 39.1 & 16.1M & 35.4 & 55.0 & 37.6  \\
        ~ & AdaptFormer~\cite{chen2022adaptformer} & 
        66.8M & 45.1 & 65.3 & 48.6 & 39.2 & 62.4 & 41.5 & 16.7M & 38.4 & 58.9 & 40.9   \\
        ~ & SSF~\cite{lian2022scaling} & 
        65.6M & 44.2 & 64.2 & 47.8 & 38.6 & 61.0 & 41.0 & 15.8M & 37.8 & 57.8 & 40.4   \\
        ~ & LoRA~\cite{hu2021lora} & 
        66.2M & 46.9 & 67.3 & 50.6 & 40.8 & 64.3 & 43.4 & 16.2M & 41.1 & 62.1 & 44.1   \\
        ~ & HST \textbf{(ours)} & 
        68.4M & \bf 49.5 & \bf 69.0 & \bf 53.9 & \bf 43.0 & \bf 66.1 & \bf 46.8 & 18.5M & \bf 46.0 & \bf 65.7 & \bf 49.7   \\
        \bottomrule
    \end{tabular}}
    \vspace{-2mm}
     \caption{Performance comparisons on object detection and instance segmentation. AP$^\text{b}$ and AP$^\text{m}$ represent box AP and mask AP, respectively. ``MS" means multi-scale training.
	}
    \label{tab:results_detection_cacade}
\vspace{-2mm}
\end{table*}

\subsection{Object Detection and Instance Segmentation}\label{det}
As shown in Table~\ref{tab:results_detection_cacade}, regardless of the detector used, existing PETL methods still exhibit a significant performance gap compared to the full-tuning. This disparity stems from the fundamental differences between classification tasks and dense prediction tasks, highlighting the ineffectiveness of existing PETL techniques in transfer learning for the latter. However, our HST breaks through this performance limit.
When training Mask R-CNN with 3$\times$ schedule, our HST demonstrates only 1.2 $\rm AP^b$ decrease and achieves equal performance in $\rm AP^m$ compared to full-tuning. Additionally, HST yields a 0.8 $\rm AP^b$ and 0.8 $\rm AP^m$ improvement over full fine-tuning in Cascade Mask R-CNN with 3$\times$ schedule, while only exhibiting a 0.7 $\rm AP^b$ decrease compared to full-tuning method in ATSS. These encouraging results indicate that our method enhances transfer robustness and even enables ViT models to achieve superior performance.
Moreover, we can observe that HST performs more satisfactorily when using larger models like ViT-L. There is a performance gap of 2.8 $\rm AP^b$ between HST and full finetune on the base model, while achieving comparative performance on the large model.

\vspace{-2mm}
\subsection{Semantic Segmentation}\label{seg}
In Table~\ref{tab:ade}, we present semantic segmentation results in terms of mIoU on ViT/B, utilizing multi-scale (MS) techniques for comparison. 
Our HST method exhibits impressive performance, achieving mIoU scores of 47.0 and 47.5 with MS when integrated with UperNet, outperforming other PETL methods by at least 2.1 mIoU while maintaining the fewest trainable parameters. Moreover, within Semantic FPN, HST attains state-of-the-art results with mIoU scores of 44.3 and 45.0 with MS. Despite these achievements, the results highlight that there is still potential for improvement in segmentation tasks compared to full fine-tuning, indicating both the ongoing challenges and the opportunities for further advancement in PETL for dense prediction tasks.

\begin{table}[t]\small
    \vspace{-2mm}
    \centering
    \renewcommand{\arraystretch}{0.93}
	\setlength\tabcolsep{1.41mm}{
	\begin{tabular}{l|c|ccc|ccc}
		\toprule
	    \multirow{2}{*}{Method}  &   \multirow{2}{*}{Crop Size} & \multicolumn{3}{c|}{Semantic FPN 80k}   & \multicolumn{3}{c}{UperNet 160k}  \\
	   & & \#Param & mIoU & +MS  & \#Param & mIoU & +MS \\
	   % & (G) & (M) & (\%) & (G) & (M) & (\%) \\
		\midrule
        \rowcolor{gray!15} 
        Full fine-tuning & 512$\times$512 & 97.7M & 46.0 & 47.2 & 127.0M & 49.5 & 50.8 \\
        Linear probing & 512$\times$512 & 11.9M & 34.2 & 36.5 & 41.2M & 37.1 & 39.1 \\
        \midrule
        VPT-deep~\citep{jia2022visual} & 512$\times$512 & 12.5M & 41.5 & 41.4 & 41.8M & 44.0 & 46.1 \\
        AdaptFormer~\citep{chen2022adaptformer} & 512$\times$512 & 13.1M & 42.8 & 43.0 & 42.4M & 43.4 & 44.6 \\
        SSF~\citep{lian2022scaling} & 512$\times$512 & 12.1M & 44.2 & 44.6 & 41.4M & 44.9 & 46.8 \\
        LoRA~\citep{hu2021lora} & 512$\times$512 & 12.5M & 44.0 & 44.9 & 41.8M & 44.9 & 46.4 \\
        HST \textbf{(ours)} & 512$\times$512 & 14.7M & \textbf{44.3} & \textbf{45.0} & 39.9M & \textbf{47.0} & \textbf{47.5} \\
        
	\bottomrule
	\end{tabular}}
        \vspace{1mm}
	\caption{Semantic segmentation on the ADE20K val.``MS" means multi-scale testing.}
	\label{tab:ade}
 \vspace{-4mm}
\end{table}

\vspace{-2mm}
\subsection{Efficiency Analysis}\label{efficiency}
To demonstrate the inference and training efficiency of our method, we provide a detailed efficiency analysis of HST in Appendix~\ref{appendix:eff_analysis}.

\subsection{Visualizations}\label{vis}
\begin{figure}[h]
    \centering
    \includegraphics[width=1.0\linewidth]{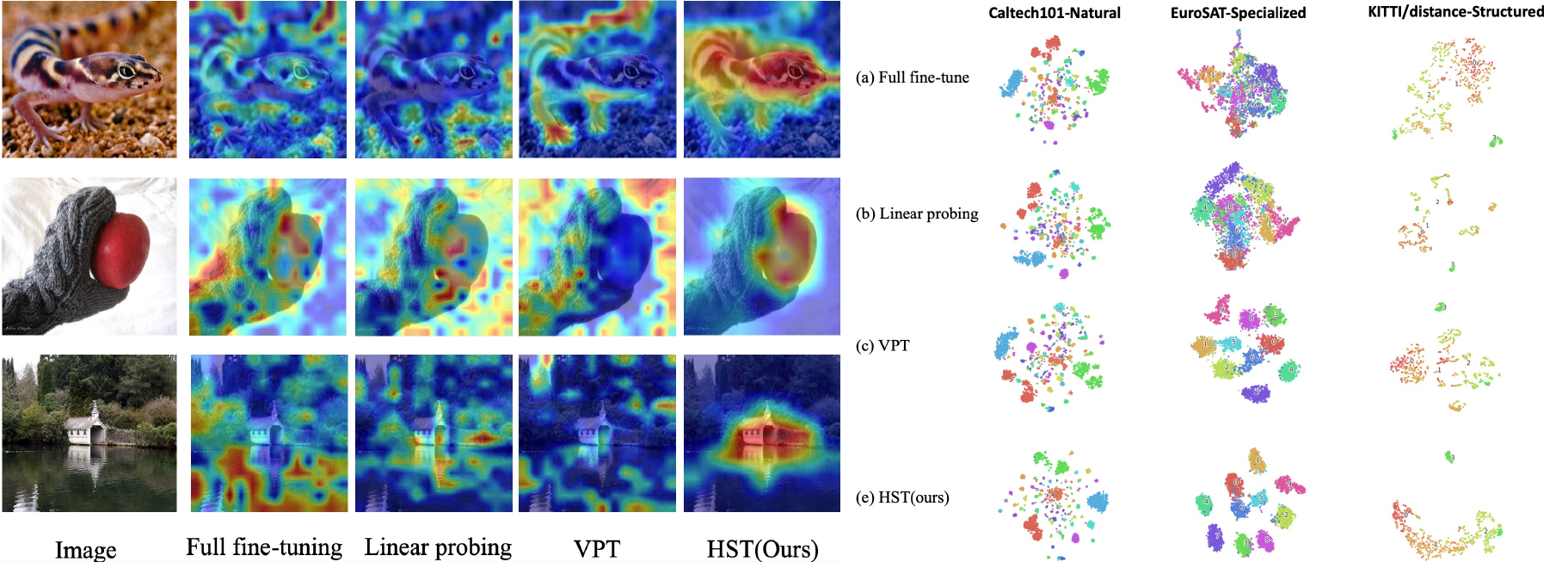}
    \vspace{1mm}
    \caption{
    \textbf{Left: }Visualization of attention maps. \textbf{Right: }t-SNE visualization of various PETL methods applied to three tasks within different categories.
    }
    \label{fig:vis}
\end{figure}
As illustrated in Figure~\ref{fig:vis}, we employ t-SNE~\cite{van2008visualizing} to visualize the feature distributions of HST and other PETL methods, revealing that HST significantly enhances feature clustering. Furthermore, we use Grad-CAM~\cite{selvaraju2017grad} to visualize attention maps, demonstrating that HST distinctly highlights target objects. This capability underlines why HST excels in dense prediction tasks—its adeptness at grounding the main object, supported by HSN's effective modeling of multi-scale features. (Additional visualizations can be found in the Appendix~\ref{appendix:vis}.)

\subsection{Ablation Studies}\label{abl}
We conducted an ablation study on HST to identify key factors influencing its effectiveness, using the VTAB-1K validation set and MS COCO with the Mask R-CNN 1$\times$ schedule for all tests.

\vspace{-1mm}
\paragraph{Number of Meta-Register}
Table~\ref{abl:num_mr} illustrates the impact of adjusting the number of trainable tokens in Meta-Register tuning performance. The quantity of Meta-Register within HST is crucial in determining computational complexity. Unlike the observations in VPT, increasing the number of trainable tokens in HST does not yield significant performance enhancements. Instead, using just one trainable token is enough to achieve satisfactory results in classification transfer tasks. While using 32 trainable tokens offers a marginal improvement, it substantially raises both training and inference costs. Furthermore, we found that a higher count, such as 64 tokens, actually diminishes performance in classification tasks and nearly reaches performance saturation in dense prediction tasks. Therefore, to strike a balance between speed and accuracy, it is advisable to select one trainable token in Meta-Register.

\noindent
\begin{table}[ht]
\vspace{-6mm}
\centering
\begin{minipage}{0.4\textwidth}
    \centering
    \renewcommand{\arraystretch}{0.9}
    \setlength\tabcolsep{2.0mm}
    \begin{tabular}{c|c|cc}
        \toprule
        \multirow{1}{*}{$N$} 
        & Mean(\%)
        & $\rm AP^b$ & $\rm AP^m$  \\
        \midrule
        1  & 76.1  & 40.3 & 38.0   \\
        32  & 76.2  & 40.4 & 38.2   \\
        64 & 75.9  & 40.5 & 38.2   \\
        \bottomrule
    \end{tabular}
    \vspace{1mm}
    \caption{Number of Meta-Register.}
    \label{abl:num_mr}
\end{minipage}\hfill
\begin{minipage}{0.55\textwidth}
    \centering
    \renewcommand{\arraystretch}{0.95}
    \setlength\tabcolsep{1.0mm}
    \begin{tabular}{c|c|cc}
        \toprule
        \multirow{1}{*}{Method} 
        & Mean(\%)
        & $\rm AP^b$ & $\rm AP^m$  \\
        \midrule
        only GlobalT  & 75.3  & 38.7 & 36.5   \\
        only Meta-Register  & 75.7  & 39.5 & 37.3   \\
        Meta-Register + GlobalT  & 76.1  & 40.3 & 38.0   \\
        \bottomrule
    \end{tabular}
    \vspace{1mm}
    \caption{The effect of Meta-Register.}
    \label{abl:eff_mr}
\end{minipage}
\vspace{-6mm}
\end{table}

\vspace{-1mm}
\paragraph{Effect of Meta-Register}
As shown in Table~\ref{abl:eff_mr}, the performance achieved using only GlobalT does not exceed that obtained with Meta-Register alone. This outcome is primarily due to the Meta-Register’s ability to adaptively extract more enriched global features from each ViT block. However, when combined, they achieve optimal performance.

\begin{table*}[h]\small
    \renewcommand{\arraystretch}{0.95}
    \centering
    \setlength\tabcolsep{0.8mm}
    \begin{tabular}{c|cccc|cccc}
        \toprule
         \multirow{2}{*}{Method} & \multicolumn{4}{c|}{Components} & \multirow{2}{*}{\#Param} & \multirow{2}{*}{Mean(\%)} & \multirow{2}{*}{${\rm AP^b}$} & \multirow{2}{*}{${\rm AP^b}$}  \\
          & LN-Tuning & Weight-Sharing & GlobalT & FG Injection \\
	\midrule
         ViT-B w/. HSN  & ~ & ~ & ~ & ~  & 1.07M & 72.1 & 30.0 & 29.2  \\ 
         HST.a & \checkmark & ~ & ~  &  ~  & 1.10M & 74.3 & 32.8 & 31.5  \\ 
         HST.b & \checkmark  & \checkmark & ~ & ~  & 0.78M & 75.0 & 32.8 & 31.5  \\ 
         HST.c & \checkmark  & \checkmark & \checkmark & ~  & 0.78M & 75.2 & 34.8 & 33.6   \\ 
         HST.d & \checkmark  & \checkmark & ~ & \checkmark & 0.78M & 75.7 & 39.5 & 37.3  \\ 
         HST \textbf{(ours)} & \checkmark  & \checkmark & \checkmark & \checkmark   & 0.78M  & \textbf{76.1} & \textbf{40.3} & \textbf{38.0}  \\ 
	\bottomrule
    \end{tabular}
    \vspace{-2mm}
    \caption{Ablation studies of key components}
    \label{abl:component}
    \vspace{-2mm}
\end{table*}

\paragraph{Ablation for Components}
To explore the impact of each key design element, we progressively enhance ViT-B with HSN to develop the final version of HST. As detailed in Table~\ref{abl:component}, training HSN alone achieves a baseline accuracy of 72.1\% on VTAB-1K and scores of 30.0 $\rm AP^b$ and 29.2 $\rm AP^m$ on MSCOCO. With the addition of the LN tuning method, the \textbf{HST.a} model shows improvements of 2.2\%, 2.8 $\rm AP^b$, and 2.3 $\rm AP^m$ over the baseline. In \textbf{HST.b}, we discover that linear weight sharing surpasses the performance of multiple linear layers, likely due to implicit feature fusion provided by the shared layers. Moreover, by integrating 'GlobalT' with Meta-Register as an injection in the Side block, \textbf{HST.c} achieves further gains of 0.2\% in classification accuracy, 2.0 for $\rm AP^b$, and 2.1 for $\rm AP^m$. Additionally, a separate experiment (\textbf{HST.d}) focusing solely on Fine-Grained (FG) Injection without GlobalT yielded significant performance enhancements. Ultimately, implementing all proposed components together in HST led to substantial overall improvements of 4.0\% in classification accuracy, 10.3 for $\rm AP^b$, and 8.8 for $\rm AP^m$, confirming the significance of each component.

\section{Conclusion}
In this paper, we introduce Hierarchical Side-Tuning (HST), a new parameter-efficient transfer learning method designed to effectively adapt large vision Transformer backbones.
Our tuning framework incorporates a trainable hierarchical side network, which successfully leverages the intermediate features of the pre-trained model and generates multi-scale features for making predictions. 
Extensive experiments illustrate that HST consistently outperforms previous state-of-the-art methods on diverse benchmarks, significantly reducing the performance disparity between PETL methods and full fine-tuning in dense prediction tasks.
We hope that HST will inspire researchers into developing versatile PETL techniques applicable to a wide range of downstream tasks.

\clearpage
{\small
\bibliographystyle{plain}
\bibliography{ref}

\begin{thebibliography}{10}

\bibitem{bao2021beit}
Hangbo Bao, Li~Dong, Songhao Piao, and Furu Wei.
\newblock Beit: Bert pre-training of image transformers.
\newblock {\em arXiv preprint arXiv:2106.08254}, 2021.

\bibitem{brown2020language}
Tom Brown, Benjamin Mann, Nick Ryder, Melanie Subbiah, Jared~D Kaplan, Prafulla Dhariwal, Arvind Neelakantan, Pranav Shyam, Girish Sastry, Amanda Askell, et~al.
\newblock Language models are few-shot learners.
\newblock {\em Advances in neural information processing systems}, 33:1877--1901, 2020.

\bibitem{cai2019cascade}
Zhaowei Cai and Nuno Vasconcelos.
\newblock Cascade r-cnn: High quality object detection and instance segmentation.
\newblock {\em IEEE transactions on pattern analysis and machine intelligence}, 43(5):1483--1498, 2019.

\bibitem{caron2021emerging}
Mathilde Caron, Hugo Touvron, Ishan Misra, Herv{\'e} J{\'e}gou, Julien Mairal, Piotr Bojanowski, and Armand Joulin.
\newblock Emerging properties in self-supervised vision transformers.
\newblock In {\em Proceedings of the IEEE/CVF international conference on computer vision}, pages 9650--9660, 2021.

\bibitem{chen2019mmdetection}
Kai Chen, Jiaqi Wang, Jiangmiao Pang, Yuhang Cao, Yu~Xiong, Xiaoxiao Li, Shuyang Sun, Wansen Feng, Ziwei Liu, Jiarui Xu, et~al.
\newblock Mmdetection: Open mmlab detection toolbox and benchmark.
\newblock {\em arXiv preprint arXiv:1906.07155}, 2019.

\bibitem{chen2022adaptformer}
Shoufa Chen, Chongjian Ge, Zhan Tong, Jiangliu Wang, Yibing Song, Jue Wang, and Ping Luo.
\newblock Adaptformer: Adapting vision transformers for scalable visual recognition.
\newblock {\em Advances in Neural Information Processing Systems}, 35:16664--16678, 2022.

\bibitem{chen2021simple}
Wuyang Chen, Xianzhi Du, Fan Yang, Lucas Beyer, Xiaohua Zhai, Tsung-Yi Lin, Huizhong Chen, Jing Li, Xiaodan Song, Zhangyang Wang, et~al.
\newblock A simple single-scale vision transformer for object localization and instance segmentation.
\newblock {\em arXiv preprint arXiv:2112.09747}, 2021.

\bibitem{vit_adapter}
Zhe Chen, Yuchen Duan, Wenhai Wang, Junjun He, Tong Lu, Jifeng Dai, and Yu~Qiao.
\newblock Vision transformer adapter for dense predictions.
\newblock {\em arXiv preprint arXiv:2205.08534}, 2022.

\bibitem{mmseg2020}
MMSegmentation Contributors.
\newblock {MMSegmentation}: Openmmlab semantic segmentation toolbox and benchmark.
\newblock \url{https://github.com/open-mmlab/mmsegmentation}, 2020.

\bibitem{deng2009imagenet}
Jia Deng, Wei Dong, Richard Socher, Li-Jia Li, Kai Li, and Li~Fei-Fei.
\newblock Imagenet: A large-scale hierarchical image database.
\newblock In {\em 2009 IEEE conference on computer vision and pattern recognition}, pages 248--255. Ieee, 2009.

\bibitem{devlin2018bert}
Jacob Devlin, Ming-Wei Chang, Kenton Lee, and Kristina Toutanova.
\newblock Bert: Pre-training of deep bidirectional transformers for language understanding.
\newblock {\em arXiv preprint arXiv:1810.04805}, 2018.

\bibitem{vit}
Alexey Dosovitskiy, Lucas Beyer, Alexander Kolesnikov, Dirk Weissenborn, Xiaohua Zhai, Thomas Unterthiner, Mostafa Dehghani, Matthias Minderer, Georg Heigold, Sylvain Gelly, et~al.
\newblock An image is worth 16x16 words: Transformers for image recognition at scale.
\newblock In {\em International Conference on Learning Representations}.

\bibitem{gao2022convmae}
Peng Gao, Teli Ma, Hongsheng Li, Ziyi Lin, Jifeng Dai, and Yu~Qiao.
\newblock Convmae: Masked convolution meets masked autoencoders.
\newblock {\em arXiv preprint arXiv:2205.03892}, 2022.

\bibitem{gebru2017fine}
Timnit Gebru, Jonathan Krause, Yilun Wang, Duyun Chen, Jia Deng, and Li~Fei-Fei.
\newblock Fine-grained car detection for visual census estimation.
\newblock In {\em Proceedings of the AAAI Conference on Artificial Intelligence}, volume~31, 2017.

\bibitem{he2021towards}
Junxian He, Chunting Zhou, Xuezhe Ma, Taylor Berg-Kirkpatrick, and Graham Neubig.
\newblock Towards a unified view of parameter-efficient transfer learning.
\newblock In {\em International Conference on Learning Representations}, 2021.

\bibitem{he2022masked}
Kaiming He, Xinlei Chen, Saining Xie, Yanghao Li, Piotr Doll{\'a}r, and Ross Girshick.
\newblock Masked autoencoders are scalable vision learners.
\newblock In {\em Proceedings of the IEEE/CVF conference on computer vision and pattern recognition}, pages 16000--16009, 2022.

\bibitem{he2017mask}
Kaiming He, Georgia Gkioxari, Piotr Doll{\'a}r, and Ross Girshick.
\newblock Mask r-cnn.
\newblock In {\em Proceedings of the IEEE international conference on computer vision}, pages 2961--2969, 2017.

\bibitem{he2022hyperprompt}
Yun He, Steven Zheng, Yi~Tay, Jai Gupta, Yu~Du, Vamsi Aribandi, Zhe Zhao, YaGuang Li, Zhao Chen, Donald Metzler, et~al.
\newblock Hyperprompt: Prompt-based task-conditioning of transformers.
\newblock In {\em International Conference on Machine Learning}, pages 8678--8690. PMLR, 2022.

\bibitem{houlsby2019parameter}
Neil Houlsby, Andrei Giurgiu, Stanislaw Jastrzebski, Bruna Morrone, Quentin De~Laroussilhe, Andrea Gesmundo, Mona Attariyan, and Sylvain Gelly.
\newblock Parameter-efficient transfer learning for nlp.
\newblock In {\em International Conference on Machine Learning}, pages 2790--2799. PMLR, 2019.

\bibitem{hu2021lora}
Edward~J Hu, Yelong Shen, Phillip Wallis, Zeyuan Allen-Zhu, Yuanzhi Li, Shean Wang, Lu~Wang, and Weizhu Chen.
\newblock Lora: Low-rank adaptation of large language models.
\newblock {\em arXiv preprint arXiv:2106.09685}, 2021.

\bibitem{jia2022visual}
Menglin Jia, Luming Tang, Bor-Chun Chen, Claire Cardie, Serge Belongie, Bharath Hariharan, and Ser-Nam Lim.
\newblock Visual prompt tuning.
\newblock In {\em European Conference on Computer Vision}, pages 709--727. Springer, 2022.

\bibitem{khosla2011novel}
Aditya Khosla, Nityananda Jayadevaprakash, Bangpeng Yao, and Fei-Fei Li.
\newblock Novel dataset for fine-grained image categorization: Stanford dogs.
\newblock In {\em Proc. CVPR workshop on fine-grained visual categorization (FGVC)}, volume~2. Citeseer, 2011.

\bibitem{kirillov2019panoptic}
Alexander Kirillov, Ross Girshick, Kaiming He, and Piotr Doll{\'a}r.
\newblock Panoptic feature pyramid networks.
\newblock In {\em CVPR}, pages 6399--6408, 2019.

\bibitem{kirillov2023segment}
Alexander Kirillov, Eric Mintun, Nikhila Ravi, Hanzi Mao, Chloe Rolland, Laura Gustafson, Tete Xiao, Spencer Whitehead, Alexander~C Berg, Wan-Yen Lo, et~al.
\newblock Segment anything.
\newblock {\em arXiv preprint arXiv:2304.02643}, 2023.

\bibitem{krizhevsky2009learning}
Alex Krizhevsky, Geoffrey Hinton, et~al.
\newblock Learning multiple layers of features from tiny images.
\newblock 2009.

\bibitem{lester2021power}
Brian Lester, Rami Al-Rfou, and Noah Constant.
\newblock The power of scale for parameter-efficient prompt tuning.
\newblock {\em arXiv preprint arXiv:2104.08691}, 2021.

\bibitem{li2021prefix}
Xiang~Lisa Li and Percy Liang.
\newblock Prefix-tuning: Optimizing continuous prompts for generation.
\newblock {\em arXiv preprint arXiv:2101.00190}, 2021.

\bibitem{li2022exploring}
Yanghao Li, Hanzi Mao, Ross Girshick, and Kaiming He.
\newblock Exploring plain vision transformer backbones for object detection.
\newblock In {\em European Conference on Computer Vision}, pages 280--296. Springer, 2022.

\bibitem{li2021benchmarking}
Yanghao Li, Saining Xie, Xinlei Chen, Piotr Dollar, Kaiming He, and Ross Girshick.
\newblock Benchmarking detection transfer learning with vision transformers.
\newblock {\em arXiv preprint arXiv:2111.11429}, 2021.

\bibitem{lian2022scaling}
Dongze Lian, Daquan Zhou, Jiashi Feng, and Xinchao Wang.
\newblock Scaling \& shifting your features: A new baseline for efficient model tuning.
\newblock {\em Advances in Neural Information Processing Systems}, 35:109--123, 2022.

\bibitem{lin2017feature}
Tsung-Yi Lin, Piotr Doll{\'a}r, Ross Girshick, Kaiming He, Bharath Hariharan, and Serge Belongie.
\newblock Feature pyramid networks for object detection.
\newblock In {\em Proceedings of the IEEE conference on computer vision and pattern recognition}, pages 2117--2125, 2017.

\bibitem{lin2014microsoft}
Tsung-Yi Lin, Michael Maire, Serge Belongie, James Hays, Pietro Perona, Deva Ramanan, Piotr Doll{\'a}r, and C~Lawrence Zitnick.
\newblock Microsoft coco: Common objects in context.
\newblock In {\em Computer Vision--ECCV 2014: 13th European Conference, Zurich, Switzerland, September 6-12, 2014, Proceedings, Part V 13}, pages 740--755. Springer, 2014.

\bibitem{liu2022p}
Xiao Liu, Kaixuan Ji, Yicheng Fu, Weng Tam, Zhengxiao Du, Zhilin Yang, and Jie Tang.
\newblock P-tuning: Prompt tuning can be comparable to fine-tuning across scales and tasks.
\newblock In {\em Proceedings of the 60th Annual Meeting of the Association for Computational Linguistics (Volume 2: Short Papers)}, pages 61--68, 2022.

\bibitem{swin}
Ze~Liu, Yutong Lin, Yue Cao, Han Hu, Yixuan Wei, Zheng Zhang, Stephen Lin, and Baining Guo.
\newblock Swin transformer: Hierarchical vision transformer using shifted windows.
\newblock In {\em Proceedings of the IEEE/CVF international conference on computer vision}, pages 10012--10022, 2021.

\bibitem{loshchilov2017decoupled}
Ilya Loshchilov and Frank Hutter.
\newblock Decoupled weight decay regularization.
\newblock {\em arXiv preprint arXiv:1711.05101}, 2017.

\bibitem{mao2021unipelt}
Yuning Mao, Lambert Mathias, Rui Hou, Amjad Almahairi, Hao Ma, Jiawei Han, Wen-tau Yih, and Madian Khabsa.
\newblock Unipelt: A unified framework for parameter-efficient language model tuning.
\newblock {\em arXiv preprint arXiv:2110.07577}, 2021.

\bibitem{nilsback2008automated}
Maria-Elena Nilsback and Andrew Zisserman.
\newblock Automated flower classification over a large number of classes.
\newblock In {\em 2008 Sixth Indian conference on computer vision, graphics \& image processing}, pages 722--729. IEEE, 2008.

\bibitem{radford2021learning}
Alec Radford, Jong~Wook Kim, Chris Hallacy, Aditya Ramesh, Gabriel Goh, Sandhini Agarwal, Girish Sastry, Amanda Askell, Pamela Mishkin, Jack Clark, et~al.
\newblock Learning transferable visual models from natural language supervision.
\newblock In {\em International conference on machine learning}, pages 8748--8763. PMLR, 2021.

\bibitem{raffel2020exploring}
Colin Raffel, Noam Shazeer, Adam Roberts, Katherine Lee, Sharan Narang, Michael Matena, Yanqi Zhou, Wei Li, and Peter~J Liu.
\newblock Exploring the limits of transfer learning with a unified text-to-text transformer.
\newblock {\em Journal of machine learning research}, 21(140):1--67, 2020.

\bibitem{selvaraju2017grad}
Ramprasaath~R Selvaraju, Michael Cogswell, Abhishek Das, Ramakrishna Vedantam, Devi Parikh, and Dhruv Batra.
\newblock Grad-cam: Visual explanations from deep networks via gradient-based localization.
\newblock In {\em Proceedings of the IEEE international conference on computer vision}, pages 618--626, 2017.

\bibitem{sung2022lst}
Yi-Lin Sung, Jaemin Cho, and Mohit Bansal.
\newblock Lst: Ladder side-tuning for parameter and memory efficient transfer learning.
\newblock {\em Advances in Neural Information Processing Systems}, 35:12991--13005, 2022.

\bibitem{van2008visualizing}
Laurens Van~der Maaten and Geoffrey Hinton.
\newblock Visualizing data using t-sne.
\newblock {\em Journal of machine learning research}, 9(11), 2008.

\bibitem{vaswani2017attention}
Ashish Vaswani, Noam Shazeer, Niki Parmar, Jakob Uszkoreit, Llion Jones, Aidan~N Gomez, {\L}ukasz Kaiser, and Illia Polosukhin.
\newblock Attention is all you need.
\newblock {\em Advances in neural information processing systems}, 30, 2017.

\bibitem{wah2011caltech}
Catherine Wah, Steve Branson, Peter Welinder, Pietro Perona, and Serge Belongie.
\newblock The caltech-ucsd birds-200-2011 dataset.
\newblock 2011.

\bibitem{pvt}
Wenhai Wang, Enze Xie, Xiang Li, Deng-Ping Fan, Kaitao Song, Ding Liang, Tong Lu, Ping Luo, and Ling Shao.
\newblock Pyramid vision transformer: A versatile backbone for dense prediction without convolutions.
\newblock In {\em Proceedings of the IEEE/CVF international conference on computer vision}, pages 568--578, 2021.

\bibitem{xiao2018unified}
Tete Xiao, Yingcheng Liu, Bolei Zhou, Yuning Jiang, and Jian Sun.
\newblock Unified perceptual parsing for scene understanding.
\newblock In {\em ECCV}, pages 418--434, 2018.

\bibitem{zaken2021bitfit}
Elad~Ben Zaken, Shauli Ravfogel, and Yoav Goldberg.
\newblock Bitfit: Simple parameter-efficient fine-tuning for transformer-based masked language-models.
\newblock {\em arXiv preprint arXiv:2106.10199}, 2021.

\bibitem{zhai2019large}
Xiaohua Zhai, Joan Puigcerver, Alexander Kolesnikov, Pierre Ruyssen, Carlos Riquelme, Mario Lucic, Josip Djolonga, Andre~Susano Pinto, Maxim Neumann, Alexey Dosovitskiy, et~al.
\newblock A large-scale study of representation learning with the visual task adaptation benchmark.
\newblock {\em arXiv preprint arXiv:1910.04867}, 2019.

\bibitem{zhang2020side}
Jeffrey~O Zhang, Alexander Sax, Amir Zamir, Leonidas Guibas, and Jitendra Malik.
\newblock Side-tuning: a baseline for network adaptation via additive side networks.
\newblock In {\em Computer Vision--ECCV 2020: 16th European Conference, Glasgow, UK, August 23--28, 2020, Proceedings, Part III 16}, pages 698--714. Springer, 2020.

\bibitem{zhang2020bridging}
Shifeng Zhang, Cheng Chi, Yongqiang Yao, Zhen Lei, and Stan~Z Li.
\newblock Bridging the gap between anchor-based and anchor-free detection via adaptive training sample selection.
\newblock In {\em Proceedings of the IEEE/CVF conference on computer vision and pattern recognition}, pages 9759--9768, 2020.

\bibitem{zhang2022hivit}
Xiaosong Zhang, Yunjie Tian, Wei Huang, Qixiang Ye, Qi~Dai, Lingxi Xie, and Qi~Tian.
\newblock Hivit: Hierarchical vision transformer meets masked image modeling.
\newblock {\em arXiv preprint arXiv:2205.14949}, 2022.

\bibitem{zhang2022neural}
Yuanhan Zhang, Kaiyang Zhou, and Ziwei Liu.
\newblock Neural prompt search.
\newblock {\em arXiv preprint arXiv:2206.04673}, 2022.

\bibitem{zheng2021rethinking}
Sixiao Zheng, Jiachen Lu, Hengshuang Zhao, Xiatian Zhu, Zekun Luo, Yabiao Wang, Yanwei Fu, Jianfeng Feng, Tao Xiang, Philip~HS Torr, et~al.
\newblock Rethinking semantic segmentation from a sequence-to-sequence perspective with transformers.
\newblock In {\em Proceedings of the IEEE/CVF conference on computer vision and pattern recognition}, pages 6881--6890, 2021.

\bibitem{zhou2017scene}
Bolei Zhou, Hang Zhao, Xavier Puig, Sanja Fidler, Adela Barriuso, and Antonio Torralba.
\newblock Scene parsing through ade20k dataset.
\newblock In {\em Proceedings of the IEEE conference on computer vision and pattern recognition}, pages 633--641, 2017.

\end{thebibliography}
}

%%%%%%%%%%%%%%%%%%%%%%%%%%%%%%%%%%%%%%%%%%%%%%%%%%%%%%%%%%%%
\clearpage
\appendix
\noindent\textbf{\Large Appendix}

\section{Motivations and Sources of Inspiration} \label{Appendix:A}
In contemporary Convolutional Neural Networks (CNNs) and Vision Transformer (ViT) networks, pyramid-style architectures have become prevalent, enhancing multi-scale features and boosting performance across various applications. Despite this, many leading pre-training approaches like those used in ImageNet-21k~\cite{deng2009imagenet}, CLIP~\cite{radford2021learning}, BEIT~\cite{bao2021beit}, MAE~\cite{he2022masked}, DINO~\cite{caron2021emerging}, and SAM~\cite{kirillov2023segment}, utilize a plain ViT architecture. This raises a crucial question: how can we efficiently adapt these plainly pre-trained ViT models for dense prediction tasks? 
Some methods, such as ConvMAE~\cite{gao2022convmae} and ITPN (HiViT)\cite{zhang2022hivit}, initially shape the network with a multi-scale structure before pre-training and then apply it to subsequent tasks. However, this strategy of training from scratch is resource-intensive. An alternative involves adapting the plain ViT architecture to generate multi-scale features similar to those of a pyramid model. ViT-Adapter\cite{vit_adapter} exemplifies this approach by using a sophisticated auxiliary network, which proves more effective than simpler upsampling or downsampling techniques. However, these methods predominantly enhance ViT’s performance through full fine-tuning—a process that is becoming increasingly challenging with larger models due to its high resource and storage demands. Consequently, our research is dedicated to improving performance through parameter-efficient fine-tuning, addressing this significant challenge.

\section{Detailed Descriptions for the Evaluation Datasets and Methods} \label{Appendix:B}

\subsection{Evaluation Methods}\label{app:evaluation_methods}
$(i)$ Full fine-tuning, where all parameters of the models are updated; $(ii)$ linear probing, where only the parameters of the task head are updated. We also compare our method with recent SOTA PETL methods.
$(iii)$ Adapter~\cite{houlsby2019parameter}, where a new adapter structure with up-projection, non-linear function, and down-projection is inserted into the transformer and only the parameters of this new module are updated; $(iv)$ Bias~\cite{zaken2021bitfit}, where all the bias terms of parameters are updated; $(v)$ VPT~\cite{jia2022visual}, where the prompts are inserted into transformers as the input tokens; $(vi)$ LoRA~\cite{hu2021lora}, adopts an optimized low-rank matrix to the multi-head attention module in the transformer layers; 
$(vii)$ AdaptFormer~\cite{chen2022adaptformer}, adopts an optimized new Adapter structure to the FFN module in the transformer layers; $(viii)$ SSF~\cite{lian2022scaling}, leverages two learnable vectors to scale and shift the feature map in each transformer operation.

\subsection{Downstream Datasets}\label{app:downstream_datasets}
\subsubsection{Image Recognition}
The VTAB-1k benchmark was introduced in~\cite{zhai2019large}, comprising a comprehensive array of 19 tasks across diverse domains. These tasks are stratified into three distinct categories: Natural, encompassing images captured through conventional camera devices; Specialized, involving images procured under specific contexts such as medical and satellite imaging; and Structured, which comprises images synthesized within controlled, simulated environments, primarily exemplified by variations in object proximity.
Each task-specific dataset contains 1000 training samples with varying number of samples per class. 
Model evaluation, in this instance, is predicated on performance metrics computed across the entire test set.
We directly resize the image to 224$\times$224, following the default settings in~\cite{zhai2019large}.

\subsubsection{Object Detection and Instance Segmentation}
Our detection experiments are based on MMDetection~\cite{chen2019mmdetection} and the MS COCO dataset~\cite{lin2014microsoft}. We use 3 mainstream detectors to evaluate our HST, including Mask R-CNN~\cite{he2017mask}, Cascade Mask R-CNN~\cite{cai2019cascade} and ATSS~\cite{zhang2020bridging}. 
Following common practices~\cite{pvt}, we employ 1$\times$ and 3$\times$ training schedules with a batch size of 16. We utilize the AdamW~\cite{loshchilov2017decoupled} optimizer with an initial learning rate of $1 \times 10^{-4}$ and a weight decay of 0.05.

\subsubsection{Semantic Segmentation}
Our semantic segmentation experiments are based on MMSegmentation~\cite{mmseg2020} and the ADE20K~\cite{zhou2017scene} dataset which has 20k and 2k images from 150 categories for training and validation. We take Semantic FPN~\cite{kirillov2019panoptic} and UperNet~\cite{xiao2018unified} as the basic frameworks.
For Semantic FPN, we adopt the same settings as PVT~\cite{pvt} and train the models for 80k iterations. As for UperNet, we adhere to the Swin Transformer's~\cite{swin} settings and train it for 160k iterations.
We employ the same approach as used in detection to endow ViT with the capability to generate multi-scale feature outputs.

\section{Efficiency Analysis}\label{appendix:eff_analysis}
To validate the efficiency of HST, we compare three main factors, which are the inference speed, training memory and training time with HST and existing PETL methods.

\paragraph{Training}
\begin{figure}[h]
    \centering
    \includegraphics[width=1.0\linewidth]{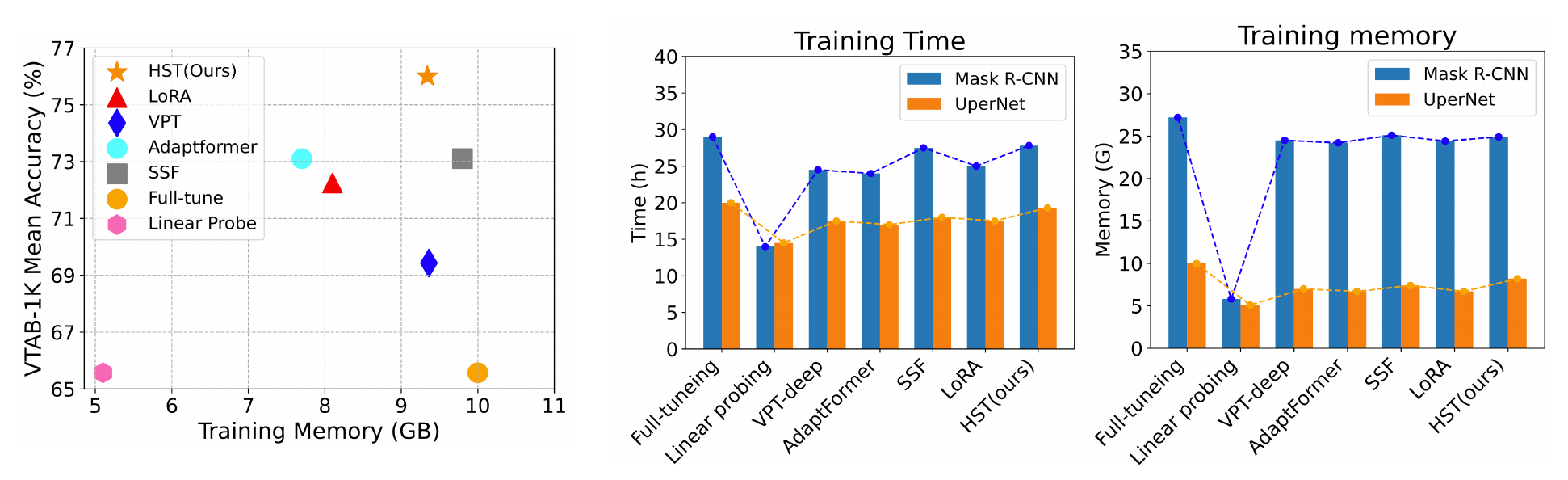}
    \vspace{-4mm}
    \caption{
    Comparative analysis of training memory and time across various visual tasks using different PETL methods.
    }
    \label{fig:training_efficiency}
\end{figure}
As illustrated in Figure~\ref{fig:training_efficiency}, our findings indicate that in the image classification benchmark, HST demands a training memory requirement similar to that of VPT (with 64 prompts), yet less than SSF and full fine-tuning methods. Remarkably, it still achieves the highest accuracy at 76.12\%. For dense prediction benchmarks, HST requires a training duration comparable to SSF, though it is slightly longer than that required by AdaptFormer and LoRA. Regarding training memory, these PETL methods demonstrate closely aligned profiles, all of which are more efficient than those required by full fine-tuning.

\paragraph{Inference}
\begin{table}[h]
    \centering
    \renewcommand{\arraystretch}{0.93}
	\setlength\tabcolsep{3mm}{
	\begin{tabular}{l|c|ccc}
		\toprule
	    \multirow{2}{*}{Method}  &   \multirow{1}{*}{FLOPs} & \multicolumn{3}{c}{GPU latency (imgs/sec)}  \\
	   & (G) & bs=1 & bs=32 & bs=128 \\
	\midrule
        % \rowcolor{gray!5} 
        Full fine-tuning & 16.9 & 118.0 & 302.8 & 306.0  \\
        \midrule
        VPT-deep & 22.3 & 116.0 & 216.5 & 229.6 \\
        AdaptFormer & 17.1 & 101.0 & 291.5 & 296.2  \\
        SSF & 16.9 & 93.4 & 269.0 & 274.5 \\
        LoRA & 17.0 & 88.6 & 290.3 & 294.2 \\
        HST (serial) & 17.5 & 70.5 & 240.2 & 248.1   \\
        HST (parallel) & 17.5 & 96.5 & 277.2 & 284.1   \\
	\bottomrule
	\end{tabular}}
    \vspace{1mm}
    \caption{\textbf{Efficiency comparison.} We use ViT-B/16 as the backbone. The inference speed is defined by images per second (imgs/sec). All results are the average of 100 trials.}
    \label{tab:eff_cls}
\end{table}
To evaluate the inference efficiency of various PETL methods, we present GPU latency in this section. In Table~\ref{tab:eff_cls}, we compare inference speeds across the classification benchmark. Notably, all PETL methods introduce varying degrees of inference slowdown. We have observed that for single-batch inference, factors such as network depth and the inclusion of additional network units significantly impact GPU latency. Conversely, in multi-batch inference, the critical factor affecting GPU latency is the number of tokens input into the Transformer. For example, employing a batch size of 1 in VPT results in latency nearly equivalent to that of full fine-tuning. However, with batch sizes of 32 or 128, latency significantly increases.
Regarding HST, the incorporation of a hierarchical side network demands greater computational resources, consequently resulting in slightly slower inference speeds compared to other PETL methods. However, our approach potentially accelerates inference speed through optimized parallel computation. Specifically, our method facilitates concurrent computation, allowing calculations in both the ViT network and the HSN to progress independently. The HSN can compute simultaneously using various ViT output features obtained during the ViT's forward process. Therefore, as shown in Table~\ref{tab:eff_cls}, by employing targeted parallel computing methods through practical engineering optimization, the inference speed of HST can be substantially enhanced.

\section{More Visualizations}\label{appendix:vis}
\subsection{Feature Quality}
We employ t-SNE to visualize the feature distributions of HST and other baseline methods, aiming to assess the quality of the generated features. These features are extracted from three distinct tasks: Caltech101, EuroSAT, and KITTI-Dist, each representing a different category. We utilize a ViT-B/16 model pretrained on the ImageNet-21K dataset as the basis for feature extraction.
In Figure~\ref{fig:vis} from main body, it is evident that both linear fine-tuning and full fine-tuning methods tend to produce mixed features. In comparison, our HST demonstrates superior feature clustering results when contrasted with VPT. This observation further validates that our HST enhances the ability of ViT to generate more distinguishable representations while requiring fewer learnable parameters.

\subsection{Attention Map}
\begin{figure}[h]
    \centering
    \includegraphics[width=1.0\linewidth]{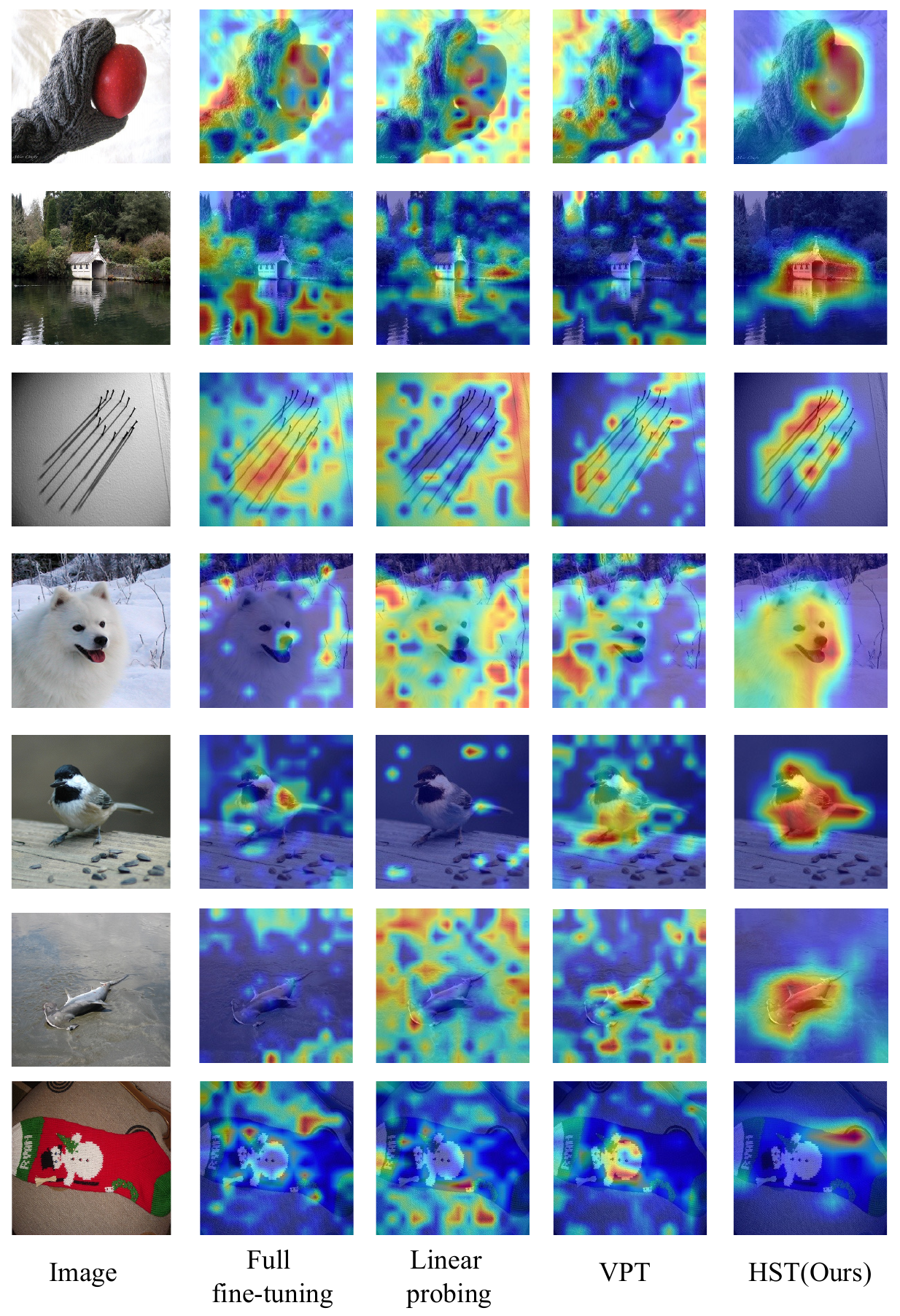}
    \vspace{-0.5em}
    \caption{\textbf{Visualization results.} We utilize Grad-CAM to visualize attention maps on the ImageNet-1k validation set. Each column presents the RGB image, full fine-tuning, linear probing, VPT-Deep, and our HST.}
    \label{fig:app:vis}
\end{figure}

We present additional attention maps from different fine-tuning methods, as illustrated in Figure~\ref{fig:app:vis}. We observe that methods such as full fine-tuning, linear probing, and VPT often demonstrate insufficient concentrated attention on the object. While effective in some images, they lack suitable attention in others. In contrast, HST consistently excels at accurately locating the intended subject of interest.

\section{Limitations and Societal Impacts}\label{appendix:E}
Regarding the limitations of this work, there are primarily two issues: (1) During the fine-tuning process, the unfreezing of parameters in the ViT's layer normalization (LN) layers results in different LN parameters for each downstream tasks. This variability hinders the simultaneous use of multiple hierarchical side networks for multi-task inference, which is a crucial functionality and future direction. (2) The segmentation experiment results suggest that the current methods of parameter-efficient fine-tuning for semantic segmentation still do not match the performance of full fine-tuning. Further research is needed to understand the underlying reasons and to explore potential solutions.

For societal impacts, our method, specifically designed for parameter-efficient fine-tuning of pre-trained models, may also inadvertently violate fine-tuning practices if the pretrained model has been trained on data obtained illegally.

\clearpage

%%%%%%%%%%%%%%%%%%%%%%%%%%%%%%%%%%%%%%%%%%%%%%%%%%%%%%%%%%%%

\end{document}